\documentclass[10pt,twocolumn,letterpaper]{article}

\usepackage[pagenumbers]{cvpr}
\usepackage{tabu}
\usepackage{tabularray}
\usepackage{multirow}
\usepackage{multicol}
\usepackage{rotating} 
\usepackage{caption}
\usepackage{subcaption}
\usepackage{pifont}
\usepackage{graphicx}
\usepackage{amsmath}
\usepackage{amssymb}
\usepackage{booktabs}
\usepackage{animate}
\usepackage{arydshln}
\usepackage{adjustbox}
\usepackage{xcolor}
\UseTblrLibrary{booktabs}
\definecolor{best}{RGB}{244,204,204}      
\definecolor{second}{RGB}{252,229,205}    

\newcommand{\method}{\textsc{Trace}\xspace}
\usepackage{amssymb} 
\usepackage{pifont}  

\newcommand{\cmark}{\ding{51}}%
\newcommand{\xmark}{\ding{55}}%

\usepackage{hyperref}

\title{\method: Object Motion Editing in Videos with First-Frame Trajectory Guidance}

\author{
Quynh Phung$^{1\ast}$ \quad Long Mai$^{2}$ \quad Cusuh Ham$^{2}$ \\
Feng Liu$^{2}$ \quad Jia-Bin Huang$^{1}$ \quad  Aniruddha Mahapatra$^{2}$ \\
\\
$^1$~University of Maryland, College Park \quad
$^2$~Adobe Research \\
{\tt\small \{quynhpt,jbhuang\}@umd.edu} \quad
{\tt\small \{malong, ham, fengl, anmahapa\}@adobe.com}\\
\url{https://trace-motion.github.io/}
}

\newcommand{\Fref}[1]{Fig.~\ref{#1}}
\newcommand{\Tref}[1]{Table~\ref{#1}}

\usepackage{pifont}

\begin{document}

\twocolumn[{%
\renewcommand\twocolumn[1][]{#1}%
\maketitle
\vspace{-30pt}
\begin{center}
    \centering
    \captionsetup{type=figure}
    \includegraphics[width=1.\textwidth]{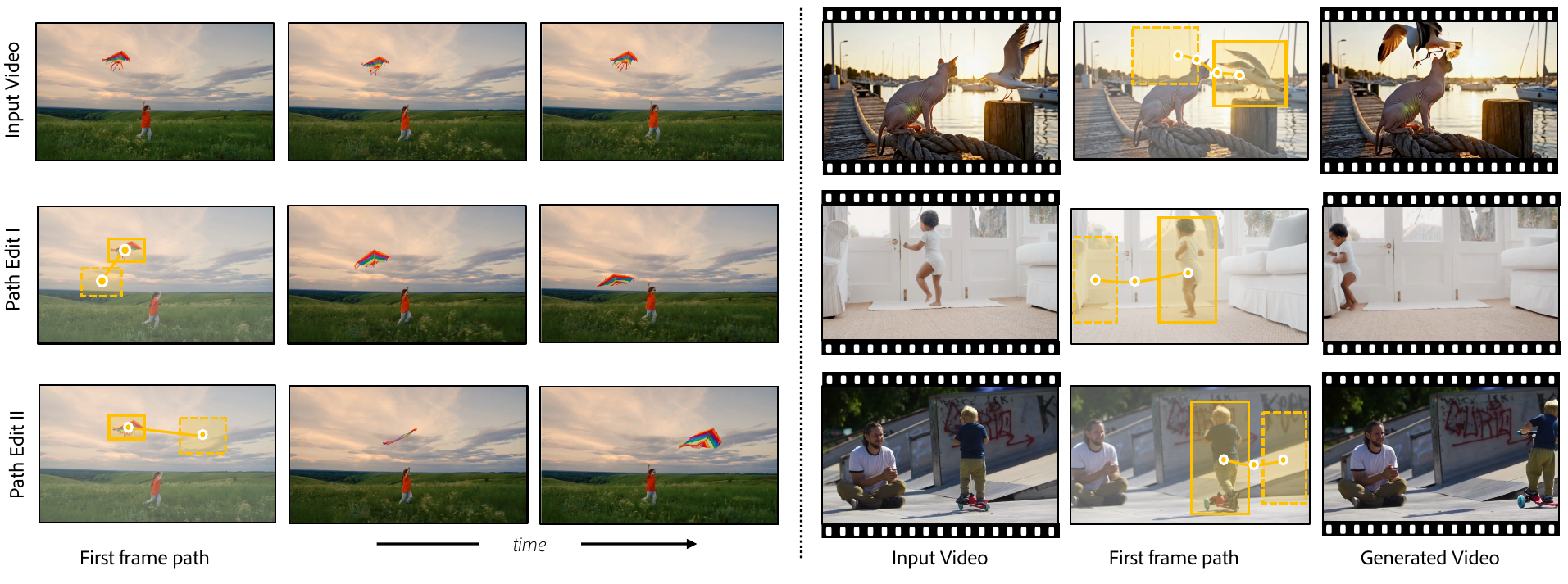}
    \caption{\textbf{\method} enables intuitive object-centric motion editing in videos.
\textbf{Left:} Given an input video (e.g., a girl flying a kite), the user specifies a desired 2D trajectory in the first-frame view using sparse bounding boxes. 
Two editing examples (Path Edit I and II) are shown. 
The solid yellow box denotes the initial bounding box, the dashed yellow box denotes the final bounding box, and the yellow curve represents the user-defined motion path. 
Our method re-synthesizes the video such that the object follows the new trajectory while preserving the remaining scene content.
\textbf{Right:} Additional examples demonstrate robust motion editing under diverse camera movements, where \method transforms the first-frame path design into temporally consistent object motion in the generated videos. Please find associated videos in Supplementary Material.}
    
    \label{fig:teaser}
\end{center}
}]

\renewcommand{\thefootnote}{$\ast$}
\footnotetext{Work was done while interns at Adobe}
\begin{abstract}
We study object motion path editing in videos, where the goal is to alter a target object's trajectory while preserving the original scene content. 
Unlike prior video editing methods that primarily manipulate appearance or rely on point-track-based trajectory control, which is often challenging for users to provide during inference, especially in videos with camera motion, we offer a practical, easy-to-use approach to controllable object-centric motion editing. 
We present \textbf{\method}, a framework that enables users to design the desired trajectory in a single anchor frame and then synthesizes a temporally consistent edited video. 
Our approach addresses this task with a two-stage pipeline: a cross-view motion transformation module that maps first-frame path design to frame-aligned box trajectories under camera motion, and a motion-conditioned video re-synthesis module that follows these trajectories to regenerate the object while preserving the remaining content of the input video. 
Experiments on diverse real-world videos show that our method produces more coherent, realistic, and controllable motion edits than recent image-to-video and video-to-video methods.
\end{abstract}    
\section{Introduction}
\label{sec:intro}

Generative video modeling has facilitated rapid advances in video content creation
\cite{makevideo,videocrafter1,videocrafter2,moviegen,lvdm,wan2025,ge2023preserve,cogvideox}. 
Modern video editing systems excel at transforming video's appearance (\eg, style transfers \cite{wang2023zero,yang2023rerender,chung2024style,bai2025uniedit} or local edits \cite{videopainter,truong2024local,gu2024via}). However, many professional workflows require the adjustment of temporal dynamics. 
For instance, a creative director editing a dynamic scene of a child running with a kite may be satisfied with the child's motion but want to alter the kite's motion, changing its speed or direction to fit different creative intents (Fig.~\ref{fig:teaser}). 
Existing tools largely lack the ability to manipulate this object-specific motion and timing while preserving the rest of the video content.



We present~\textbf{\method}, a framework for flexible editing of object motion in videos. 
The goal is to control the motion of a selected object to follow the user-specified path while meticulously preserving all other content in the input video.
We note that this problem setup requires synthesizing novel motion of the selected object rather than copying the motion from the original videos or re-rendering the original motion from different angles since the synthesized motion characteristics (\eg, action type, pacing) needs to conform to the user's specified motion path.

Existing techniques for object re-synthesis in videos typically require dense, frame-level spatiotemporal control signals, such as mask sequences or per-frame bounding boxes~\cite{vace,hunyuancus,gencompositor}. Consequently, users must consciously translate their desired scene-aware motion path into local per-frame placements. This is a tedious process that becomes especially difficult for videos with dynamic camera motion.
Inspired by the success of previous video appearance editing works that highlight the user-friendliness of applying edits to an anchor frame and automatically propagating them temporally~\cite{genprop, schneider2025neural}, we introduce a novel problem setting: \textit{Video Editing with First-Frame Object Motion Design}.
Our key idea is to allow users to define a target motion path directly on the first frame, specifying how the object would move if the camera remained fixed at that viewpoint.
This enables the use of the first frame as a scene proxy, providing an intuitive mapping between the on-screen 2D trajectory and the actual scene-space motion.
Guided by this first-frame design, our proposed framework, \method, re-synthesizes the video to move the object along the new path while preserving the original video content.

This anchor-frame path design introduces a critical technical challenge: cross-view spatial misalignment.
The user-provided motion path on the first frame naturally differs from the object’s actual screen-space position in subsequent frames due to camera dynamics, whereas modern video synthesis models typically require spatially aligned frame-level guidance.
To address this,~\method utilizes a two-stage pipeline.
First, a learnable cross-view motion transformation module predicts the corresponding per-frame bounding box sequence in the original video’s dynamic view.
Second, a conditional video re-synthesis model uses this predicted sequence to effectively erase the original motion and regenerate the object along the new path, preserving the background and other scene elements.

By enabling users to re-synthesize input videos with intuitive control over a selected object's new motion path,~\method offers new possibilities for creative applications, such as post-production motion re-planning and video object insertion with motion control.
We validate~\method's effectiveness on real-world videos with diverse content and motion-editing scenarios, demonstrating high-quality resynthesis and flexible user controls. Our experiments show that~\method can facilitate flexible manipulation of object motion that was not possible with conventional video editing models. 




In summary, our work makes the following contributions:
\begin{enumerate}
    \item We propose the problem of video editing with first-frame motion design. This novel setting enables users to re-synthesize input videos with intuitive control of a new motion path for selected objects.
    \item We present~\method, the dedicated approach to the proposed motion editing problem. 
    We develop a two-stage framework that first transforms the user-motion design specified in the first-frame view into the corresponding per-frame object placements, providing suitable control signals for the second-stage video-diffusion-based synthesis model.
	 \item We evaluate our framework on a wide range of real-world videos, demonstrating its effectiveness in producing high-quality editing results and providing flexible user controls.
\end{enumerate}


\section{Related Work}
\label{sec:related_work}

\begin{figure*}[t!]
    \centering
    \vspace{-1em}
    \includegraphics[width=0.99\textwidth]{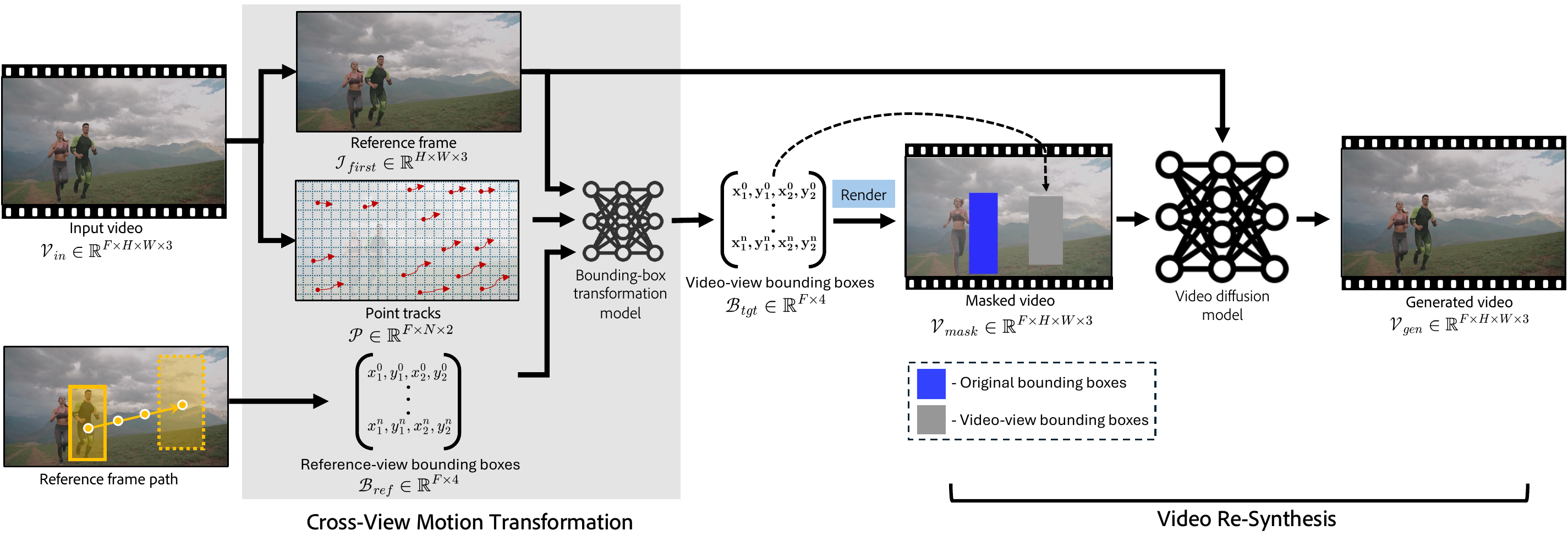}
    \vspace{-0.4em}
    \caption{ 
    \textbf{\method Overview.} 
    Our system consists of two modules. First, the \textbf{Cross-View Motion Transformation Module} converts a user's 2D input path drawn on the first-frame into a scene- and camera-aware bounding-box trajectory in the video view. Second, the \textbf{Video Re-Synthesis Module} uses this transformed box sequence to guide the generation of a new video where the object follows the desired path while inpainting the original path of the object and preserving the other parts of the input video. Both models in the figure are diffusion models.}
    
    \label{fig:overview}
\vspace{-0.5cm}
\end{figure*}

\textbf{Object motion control for video synthesis.}
Motion-controllable video generation has recently gained significant traction, especially in image-to-video (I2V) synthesis. This line of work animates a single image using control signals, ranging from text descriptions \cite{lvdm,xing2023dynamicrafter,videocrafter1,videocrafter2,moviegen,wiles2020synsin,xing2025motioncanvas} to explicit point tracks \cite{trajectoryattn,levitor,mou2024revideo,qiu2024freetraj,wang2024motionctrl}, or bounding boxes \cite{li2023trackdiffusion,huang2025fine,wang2024boximator}. While powerful for generating new content, these I2V methods are not designed for our editing task---they cannot be readily extended to handle video inputs while preserving the original camera motion, background content, and object appearance under strict constraints.

\noindent\textbf{Video editing.}
The field of video re-synthesis has a rich history, evolving from global appearance transformations~\cite{wang2023zero,yang2023rerender,chung2024style,bai2025uniedit} to fine-grained, localized edits powered by diffusion models \cite{truong2024local}. A large body of recent work focuses on training-free video editing, enabling tasks such as object attribute changes \cite{disco,qi2023fatezero,xu2025freevis} or the propagation of user edits \cite{genprop,videopainter}. The common goal of these methods is to edit appearance information (\eg, texture, style) while explicitly preserving the original motion dynamics. 
Our work addresses the orthogonal problem: we adjust the motion of a selected object while meticulously preserving its appearance and the surrounding video content.

\textit{Motion editing.}
Several concurrent works touch upon related goals, but with critical differences. Methods such as GenCompositor \cite{gencompositor} focus on generative video compositing, whereas VACE \cite{vace} learns generative priors for dynamic objects. 
These do not solve our problem of editing the path of a specific object instance within its original scene. The most relevant prior works, ReVideo~\cite{revideo}, Shape-for-Motion~\cite {shapemotion}, and Edit-by-Track \cite{lee2025generative} also reconstruct a video from a trajectory, but their capabilities are primarily demonstrated on local, small-scale motion changes or rely on explicit 3D point tracks. Furthermore, prior work on object retiming is often domain-specific (\eg, human motion) or uses NeRF-based setups, and typically only changes an object's speed along its original path, not its path itself.

\textit{First-frame editing.}
It is well-established that video editing is more intuitive when framed as a ``first-frame editing plus propagation'' task. This paradigm, used by classic methods such as NeuralAtlas~\cite{schneider2025neural} and modern generative methods such as GenProp~\cite{genprop}, frees the user from having to consistently specify edits across all frames. 
Our work is the first to adapt this intuitive paradigm to the complex task of motion editing. Achieving this conventionally would require a perfect 3D scene and camera estimation, limiting its generality. 
Instead, we introduce a dedicated cross-view motion translation module. 
This module enables first-frame motion editing, enabling our model to controllably resynthesize the video according to the new, user-defined motion path.

\section{Method}
\label{sec:method}

Our system re-synthesizes an input video using a conditional video diffusion model to move a selected object along a new, user-defined motion path, while preserving the object's appearance and all other video content.
The model generates the final video conditioned on two inputs: (1) the original video, and (2) the user-specified motion path, parameterized as a time-aligned bounding box sequence defined on the first frame of the video.

The conventional end-to-end training approach requires paired data in which each video is accompanied by the object motion from the first-frame viewpoint. Acquiring such data at scale in the real world is non-trivial, as it requires solving the challenging problem of accurately projecting the object motion from each frame onto the first-frame view. 
To address this challenge, we design a two-stage framework to decouple the problem. As illustrated in Fig.~\ref{fig:overview}, our system consists of two main components: a \textit{Cross-View Motion Transformation} module and a \textit{Video Re-Synthesis} module. 
The former translates the first-frame-based motion path that depicts the user's high-level motion intent into the corresponding boxes for each video frame, and the latter, the diffusion-based synthesis model, faithfully renders the object along this path.

\subsection{Cross-View Motion Transformation Module}
\label{sec:stage1}

\begin{figure}[t!]
    \centering
    \includegraphics[width=\linewidth]{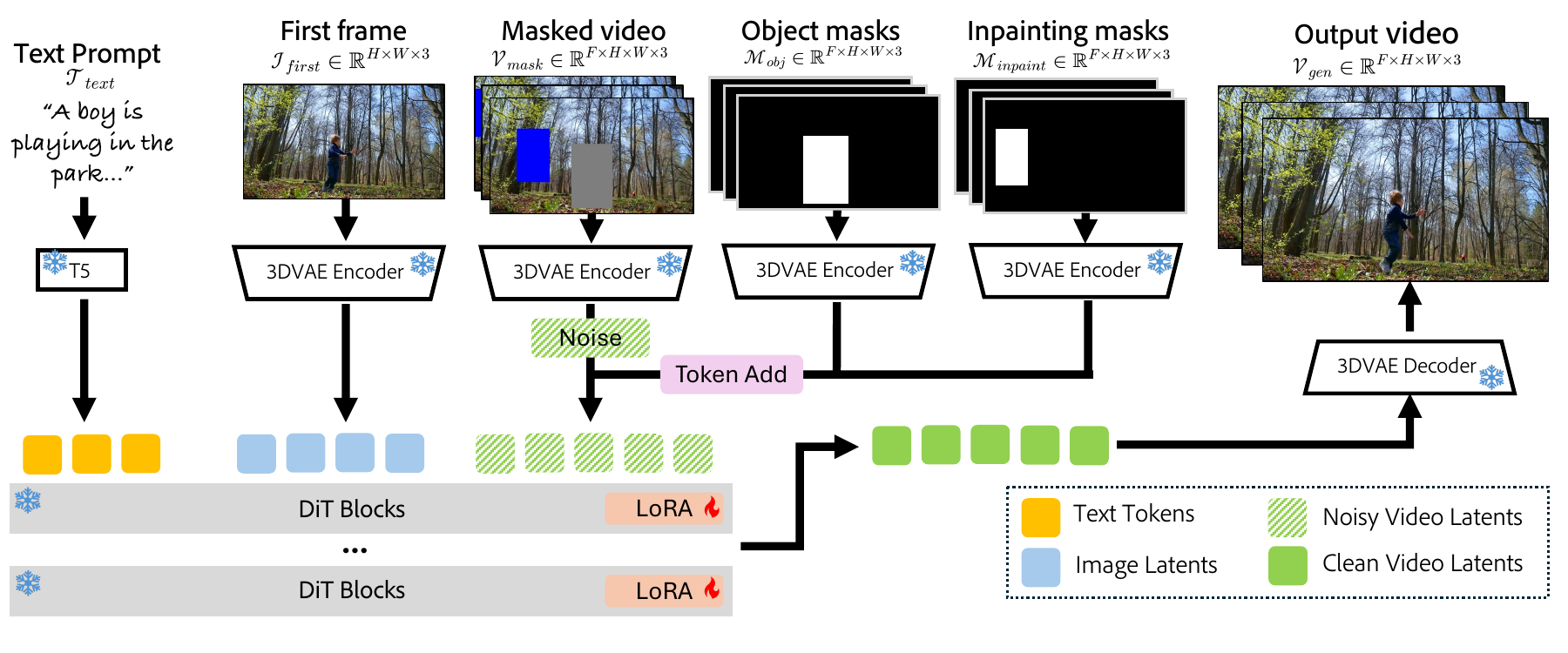}
    \vspace{-1.4em}
    \caption{\textbf{ Video Re-Synthesis Module. }
    Our model employs a Diffusion Transformer (DiT) backbone conditioned on the first frame, masked video, and binary masks (object and inpainting).
   }
    
    \label{fig:arch_stage2}
\vspace{-1em}
\end{figure}

\begin{figure*}[t!]
    \centering
    \includegraphics[width=0.95\textwidth]{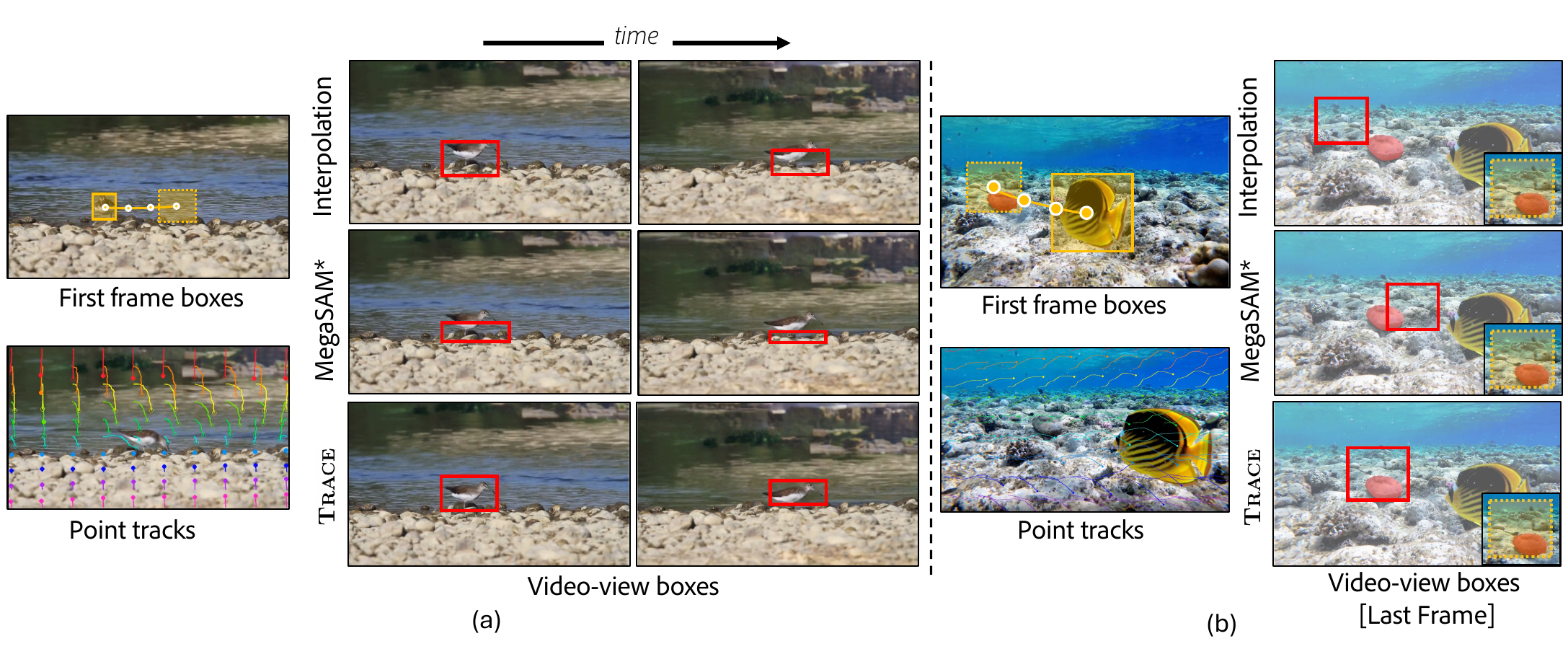}
    \vspace{-1.4em}
    \caption{
    \textbf{Cross-View Motion Transformation Comparison.} We compare our cross-view motion transformation against three baselines: (1) simple \textbf{Interpolation} of bounding boxes, (2) and 3D warping using MegaSAM \cite{megasam} to estimate depth and camera pose.\textbf{Left:} Shows the first frame path and generated video-view bounding boxes on our Cross-View Motion Transformation Module evaluation set. Only the one generated with our method \method accurately translates the first-frame view bounding boxes to the video-view. \textbf{Right:} The user's intended path moves the fish towards the static \textcolor{red}{red coral} (used as an anchor). Interpolation produces an off-track path to the right. Existing 3D warping methods yield incorrect paths due to noisy depth and pose estimation. Our cross-view transformation generates a smooth, stable, and accurate path that delivers the fish box into the coral as intended.}   
    \label{fig:stage1_baseline}
\end{figure*}

The objective of the Cross-View Motion Transformation module is to map a user-defined motion path from the static first-frame coordinate system to a temporally consistent target path that aligns with the video's dynamic camera perspective. 
As illustrated in \Fref{fig:stage1_baseline}, naive linear interpolation fails to model dynamic motion and camera changes, often producing trajectories that drift away from the intended target (e.g., last box on the right of the red coral anchor). State-of-the-art depth and pose estimators, such as Depth Anything 3 (DA3)~\cite{depthanything3} and MegaSAM~\cite{megasam}, attempt to reconstruct scene geometry, which can be used to warp the boxes, but errors in camera pose and depth estimation result in erratic, ``jumpy'' trajectories and inconsistent object scales due to accumulated noise. 
Thus, we propose to bypass explicit 3D reconstruction and treat the task as a sequence-to-sequence mapping problem, in which our model learns to implicitly model camera dynamics, yielding a smooth trajectory depicting the object's path to the user's intended destination.

\noindent\textbf{Model architecture and training.}
We use a Diffusion Transformer (DiT) consisting of eight layers to learn to predict the per-frame target bounding-box sequence $\mathcal{B}_{tgt}$ from the conditioning signal consisting of: (i) \textit{first-frame image} ($\mathcal{J}_{first}$), encoded via a 3DVAE to provide high-fidelity visual context for the static scene and camera perspective, and (ii) \textit{point trajectories} ($\mathcal{P}$), a grid of sparse point tracks that captures camera motion. Each trajectory is represented by $K=20$ discrete cosine transform (DCT) coefficients for efficiency and embedded as individual tokens within the DiT.

To handle flexible user inputs, we temporally interpolate sparse key bounding boxes into the dense box sequence $\mathcal{B}_{ref}$ on the first-frame view required by the model. 
The DiT is trained for 180k steps with a learning rate of 1.2e-4 and weight decay of 0.01 using a flow-matching objective, where the model $v_\theta$ learns to predict the velocity between pure noise $X^0$ and ground-truth tokens $X^1$:
{\small
$$\min_{\theta} \mathbb{E}_{t, X^1, X^0} \left[ \left\Vert (X^1 - X^0) - v_\theta(X^t, t \mid \mathcal{J}_{first}, \mathcal{P}, \mathcal{B}_{ref}) \right\Vert_2^2 \right].$$
}

\noindent\textbf{Data curation}. 
Collecting paired data of object boxes in the first-frame view and the corresponding boxes across a dynamic video sequence in the real world is challenging because it requires capturing the same scene with both a static camera (to define first-frame boxes) and a dynamic camera (to obtain corresponding video boxes).
Instead, we design a synthetic data pipeline that produces paired static and dynamic videos, enabling automatic construction of aligned bounding boxes for training.

First, we collect 7,500 static camera videos and extract the ``first-frame'' object bounding boxes, $\mathcal{B}_{ref}$. Next, we use ReCamMaster~\cite{recammaster} to re-render each video with 10 different dynamic camera paths and extract ``video-view'' bounding boxes $\mathcal{B}_{tgt}$, resulting in 75k synthesized videos and over 150K paired bounding box sequences. Then, we use CoTracker~\cite{karaev23cotracker} to extract a grid of $25\times25$ point trajectories for each video. Finally, we filter the data based on motion dynamics (e.g., there is a disproportionate number of static objects in static videos) and object size, yielding approximately 110k high-quality pairs. We hold out 100 videos for evaluation and use the remaining for training.

\subsection{Video Re-Synthesis Module}
\label{sec:stage2}

The goal of the Video Re-Synthesis module is to generate a new video in which a selected object follows the motion path, $\mathcal{B}_{tgt}$, generated by the previous Cross-View Motion Transformation module.
We develop a model that jointly performs \textit{object removal} (inpainting) and \textit{object generation} (synthesis), conditioned on inpainting boxes $\mathcal{M}_{inpaint}$, specifying regions to be filled with background, and synthesis boxes $\mathcal{M}_{obj}$ (derived from $\mathcal{B}_{tgt}$), specifying target regions for object generation. We mask both regions in the input video (as shown in Fig.~\ref{fig:arch_stage2}), preventing information leakage by ensuring the model cannot ``see'' the original object pixels in the inpainting region or any conflicting background in the synthesis region. Unlike in typical object insertion setups (\eg, \cite{vace, hunyuancus}) where the inserted object can be freely ``hallucinated'' without being constrained by its appearance pre-existing in the video, in this specific task, we preserve the object's fidelity in our setup using the first frame as a reference.

\noindent\textbf{Model architecture and training}.
We adapt the Wan 2.1 1.4B model \cite{wan2025}, $v_\Phi$, to predict the output video $\mathcal{V}_{gen}$, conditioned on a set of inputs $\mathbf{C}$, including a text prompt $\mathcal{T}_{text}$, first frame $\mathcal{J}_{first}$, masked video $\mathcal{V}_{mask}$, synthesis boxes $\mathcal{M}_{obj}$ and inpainting boxes $\mathcal{M}_{inpaint}$. The model is fine-tuned with LoRA \cite{hu2022lora} on 81-frame clips at a resolution of $480\times832$ and 24 fps for 8k steps. We use an AdamW optimizer with a learning rate of 1.2e-5, weight decay of 0.01, and batch size 32, and apply the flow-matching objective: 
$$\min_{\Phi} \mathbb{E}_{t, X^1, X^0} \left[ \left\Vert (X^1 - X^0) - v_\Phi(X^t, t \mid \mathbf{C}) \right\Vert_2^2 \right].$$

$v_\Phi$ iteratively uses its predicted velocity to produce a clean video latent from pure noise, which is subsequently decoded into the final pixel frames using a 3DVAE decoder.



\begin{figure*}[t]
    \centering
    \includegraphics[width=0.95\textwidth]{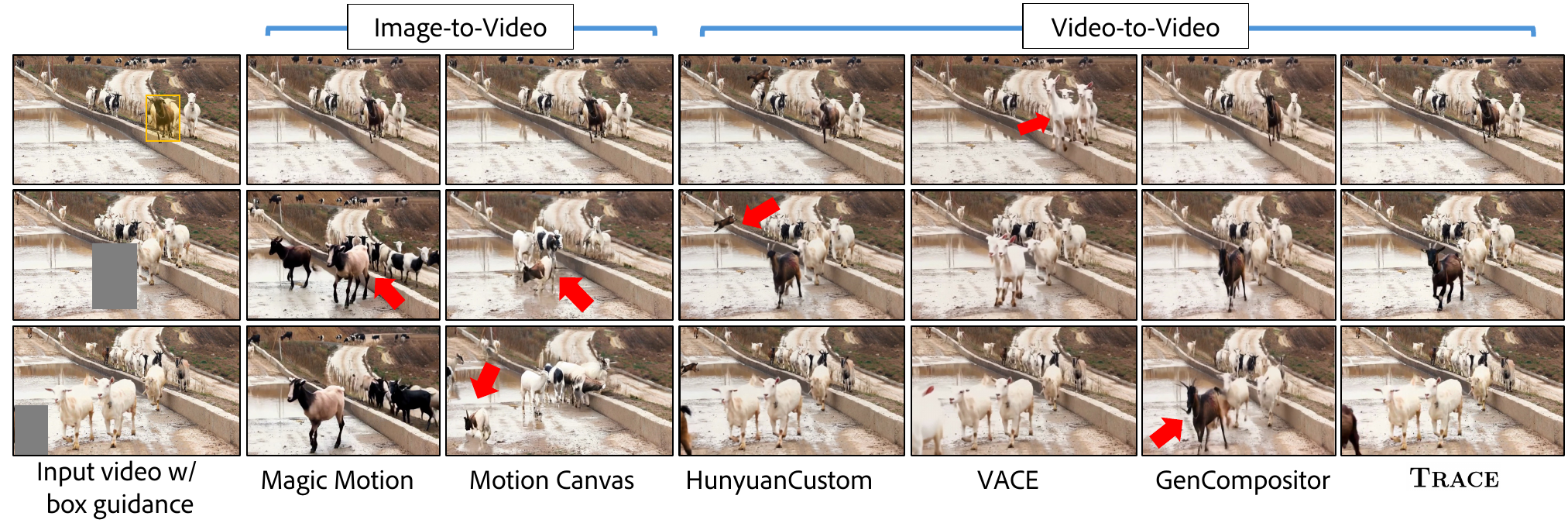}
    
    \vspace{-0.4em}
    \caption{\textbf{Comparison of Video Re-Synthesis Baselines }.Given an input video with a region (the original object) masked out and conditioned on the appearance of a reference object, the model must regenerate the object within the masked region. The goal is to produce a high-fidelity video that accurately restores the object while maintaining its original identity and temporal consistency.
   }
    
    \label{fig:baseline2}
\end{figure*}

\begin{figure*}[t]
    \centering
    \includegraphics[width=\linewidth]{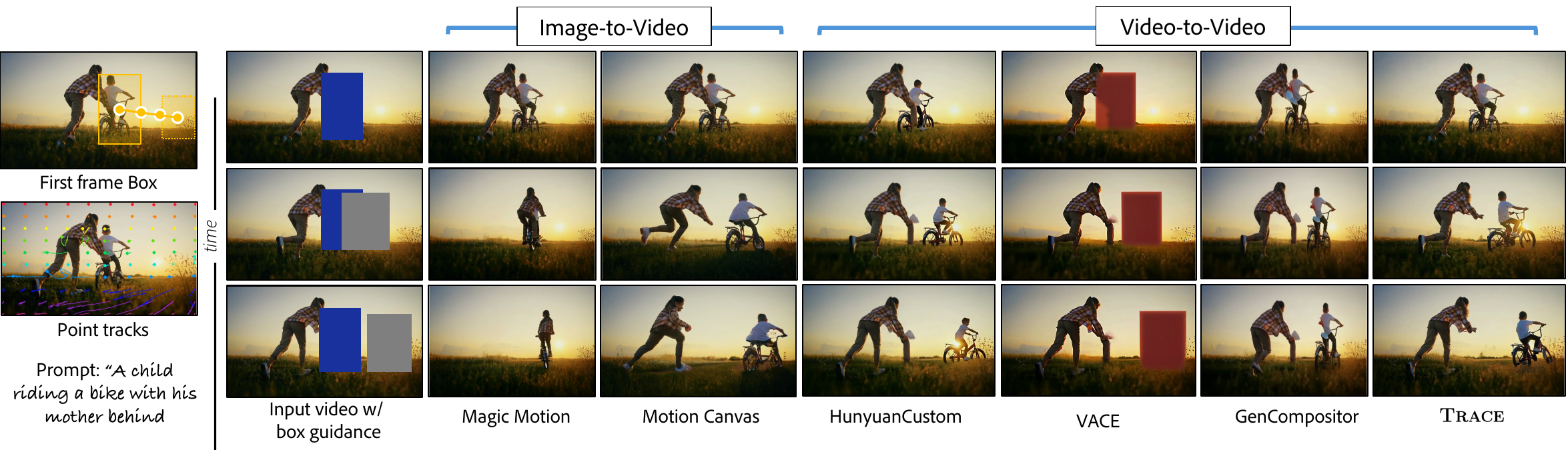}
    \vspace{-0.5em} 
    \caption{\textbf{Full pipeline comparison.} 
    Our Cross-View Motion Transformation module converts a user's path (defined by the first and last boxes in the first frame) into a sequence of video boxes. This sequence is then used to guide our re-synthesis model and three baselines (which first require a separate inpainting step). For Image-to-Video Baselines, we only use a sequence of video boxes and background point tracks (only for Motion Canvas).}
    \label{fig:baseline_full}
    \vspace{-1.0em}
\end{figure*}

\noindent\textbf{Dataset curation}. We use approximately 1.1M videos from an internal dataset for training, of which 80\% are long-shot videos, which are better suited for the motion editing task. For each video, we extract object bounding boxes using DEVA~\cite{deva}. 
To improve robustness to user-provided box guidance (rather than relying on $\mathcal{B}_{tgt}$ from the previous module), we augment the training data by smoothing bounding boxes and adding noise. We also randomly drop any of the input conditions to prevent overfitting.

\subsection {Applications}
We demonstrate that \method supports a wide range of versatile video editing tasks during inference.

\noindent \textbf{Multi-object editing}. Users can specify new trajectories for multiple objects simultaneously on the first frame, and \method processes them as parallel spatio-temporal constraints, enabling complex multi-object motion editing in a zero-shot manner (\Fref{fig:2obj}).

\begin{figure*}[t!]
    \centering
    \includegraphics[width=0.95\textwidth]{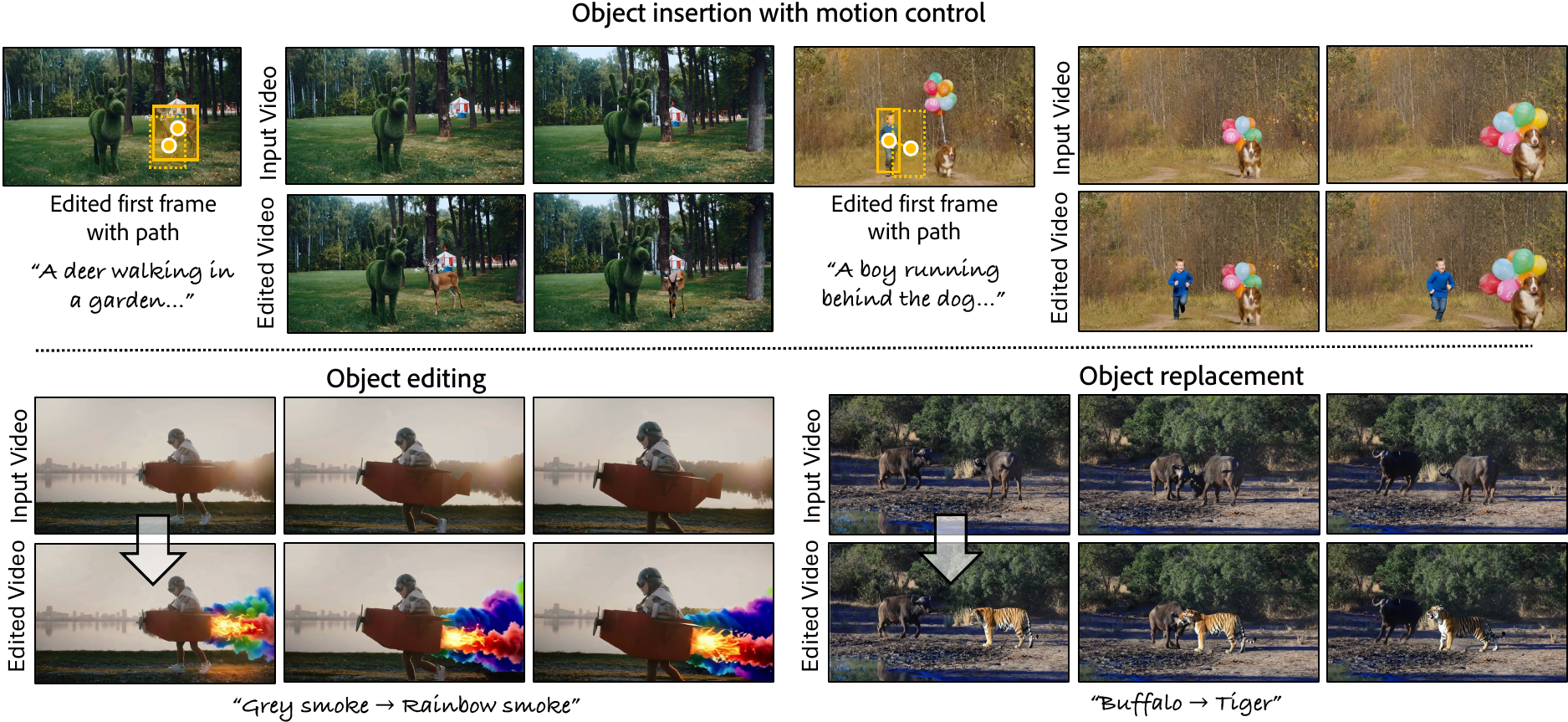}
    \vspace{-0.4em}
    \caption{\textbf{Additoinal Application. }
   Applications for diverse video editing tasks. We leverage motion flexibility for various inference-time edits: (1)\textbf{ Object insertion}: defining a first-frame layout and path, then using Qwen~\cite{qwen} to inpaint first frame and leveraging our model to propagate a new object; (2) \textbf{Object editing}: modifying first-frame attributes using Qwen and consistently propagating them along the trajectory; (3) \textbf{Object replacement}: substituting an object category in the initial frame using Qwen and re-synthesizing the sequence following original temporal dynamics.}
    
    \label{fig:application}
\vspace{-1em}
\end{figure*}

\begin{figure*}[t!]
    \centering
    \includegraphics[width=0.95\textwidth]{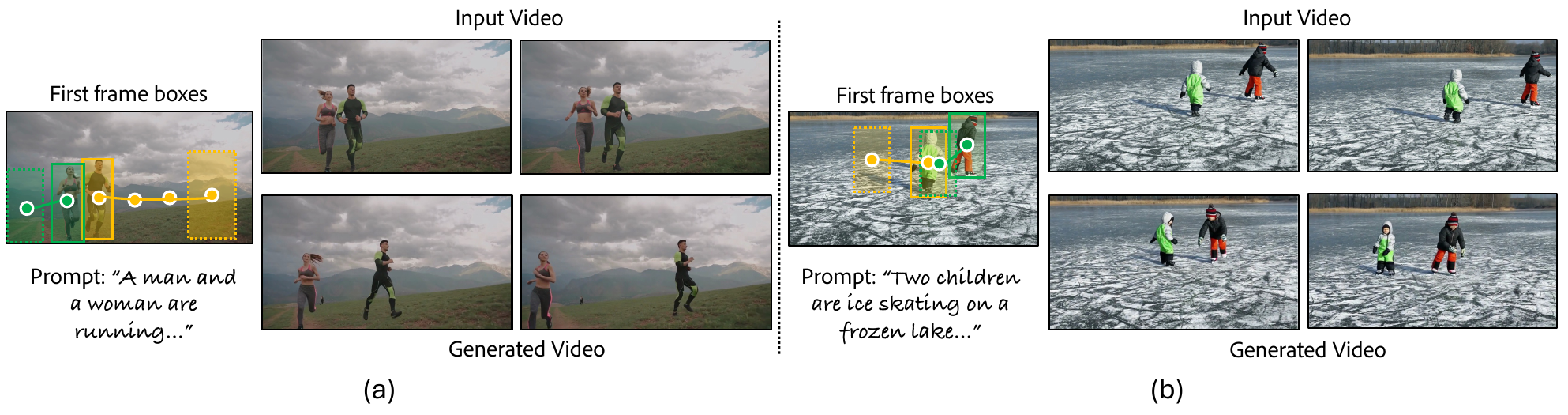}
    \vspace{-0.4em}
    \caption{\textbf{Multiple object editing. } Even though our method is trained to edit one object, it enables simultaneous zero-shot motion editing for multiple objects by processing parallel trajectories while maintaining scene consistency.
   }
    
    \label{fig:2obj}
\end{figure*}

\noindent\textbf{Object insertion}. \method enables object insertion with precise motion control, in which the user defines a target layout and path using bounding boxes in the first frame. We utilize Qwen \cite{qwen} to inpaint the new object only in the initial frame, after which our pipeline propagates the synthesized object along the designed trajectory while maintaining scene integrity (\Fref{fig:application} top).

\noindent\textbf{Object editing}. Our model can propagate localized appearance edits by modifying the object's attributes (e.g., changing its color, adding textures) in the first frame. \method ensures these edits are consistently maintained as the object follows its original path (\Fref{fig:application} bottom-left).

\noindent\textbf{Object replacement}. Users can replace a selected object with an entirely different object (e.g., replacing a deer with a tiger) in the first frame, and then \method can re-synthesize the video to ensure the new object adheres to the original temporal dynamics and motion path (\Fref{fig:application} bottom-right).

\section{Experiments}
In this section, we evaluate the effectiveness of our method in re-synthesizing the video content. We separately evaluate the effectiveness of two modules: 1) Cross-view motion transformation and 2) the Video re-synthesis.

\subsection{Cross-View Motion Transformation Module}
To evaluate the cross-view motion transformation module, we constructed an evaluation set of 100 video pairs using RecamMaster~\cite{recammaster}, following the same procedure as our training set.
The model takes bounding box sequences extracted from static-camera videos and predicts their corresponding positions in dynamic camera views.
We evaluate how well these predictions match ground-truth sequences—derived from the same videos re-rendered with synthetic camera paths—using two standard metrics: Intersection over Union (IoU)~\cite{iou} to measure spatial overlap, and mean Average Precision (mAP) at an IoU threshold of 0.5 to assess overall prediction precision.



\begin{table*}[t!]
\centering
\caption{\textbf{Full pipeline evaluation on VBench}. Performance evaluation on VBench \cite{huang2023vbench} for the full video editing pipeline. The \colorbox{best}{best} and \colorbox{second}{second-best} scores are highlighted. Higher ($\uparrow$) is better for all metrics.}
\label{tab:vbench}
\begin{adjustbox}{width=0.95\linewidth}
\begin{tblr}{
  width=\linewidth,
  colspec={@{}X[2,l] c X[1.1,c] X[1.2,c] X[1.1,c] X[1.1,c] X[1.1,c] X[1.1,c]@{}},
  row{1-Z} = {font=\small}, 
  stretch=0,
  colsep=2pt
}
\toprule
\SetCell[r=2]{l} Model & 
\SetCell[r=2]{c} Type & 
Subject & Background & Imaging & Motion & Temporal & Aesthetic \\
& & Consistency & Consistency & Quality & Smoothness & Flickering & Quality \\ 
\midrule
MagicMotion \cite{magic} & I2V & 0.9473 & 0.9571 & 0.6778 & 0.9889 & 0.7403 & 0.5878 \\
Wan-Move \cite{wan-move} & I2V & 0.9370 & 0.9542 & 0.7015 & 0.9861 & 0.6567 & \SetCell{bg=second} 0.6059 \\
Motion Canvas \cite{xing2025motioncanvas} & I2V & 0.9382 & 0.9551 & \SetCell{bg=best} 0.7152 & 0.9887 & \SetCell{bg=best} 0.8955 & 0.5978 \\
\hline[dashed]
VACE \cite{vace} & V2V + inpaint & 0.9410 & 0.9584 & 0.6494 & \SetCell{bg=second} 0.9928 & 0.6119 & 0.5777 \\
HunyuanCustom \cite{hunyuancus} & V2V + inpaint & 0.9431 & \SetCell{bg=best} 0.9609 & 0.6700 & 0.9921 & 0.6888 & \SetCell{bg=best} 0.6087 \\
GenCompositor \cite{gencompositor} & V2V + inpaint & \SetCell{bg=second} 0.9519 & 0.9471 & 0.6453 & 0.9924 & 0.5300 & 0.5503 \\
Pisco \cite{pisco} & V2V + inpaint & \SetCell{bg=best} 0.9537 & \SetCell{bg=second} 0.9593 & 0.5965 & 0.9927 & 0.7611 & 0.5416 \\
\method (Ours) & V2V & 0.9378 & 0.9376 & \SetCell{bg=second} 0.7079 & \SetCell{bg=best} 0.9994 & \SetCell{bg=second} 0.7761 & 0.5790 \\
\bottomrule
\end{tblr}
\end{adjustbox}
\end{table*}

\begin{table*}[t!]
\centering
\caption{
\textbf{Quantitative evaluation of Video Re-Synthesis Module.} 
We evaluate the video re-synthesis module on the DAVIS~\cite{davis} benchmark. 
The \colorbox{best}{best} and \colorbox{second}{second-best} scores are highlighted.
}
\label{tab:resynthesis_comparison}
\begin{adjustbox}{width=0.95\linewidth}
\begin{tblr}{
  width=\linewidth,
  colspec={@{}X[2.5,l] c c X[1,c] X[1,c] X[1,c] X[1.2,c] X[1,c]@{}},
  colsep=2pt,
  stretch=0.75,        
  rows = {rowsep=1pt}, 
  row{1-Z} = {font=\small},
}
\toprule
\SetCell[r=2]{c} Model & 
\SetCell[r=2]{c} Type & 
\SetCell[r=2]{c} Base Model & 
\SetCell[c=3]{c} Similarity and Quality & & & 
\SetCell[c=2]{c} Box Alignment \\
\cmidrule[lr]{4-6} \cmidrule[lr]{7-8}
& & & PSNR ($\uparrow$) & SSIM ($\uparrow$) & LPIPS ($\downarrow$) & Tube IoU ($\uparrow$) & mAP ($\uparrow$) \\
\midrule
MagicMotion~\cite{magic} & I2V & CogVideoX & 14.39 & 0.34 & 0.50 & 0.43 & 0.35 \\
Wan-Mover~\cite{wan-move} & I2V & Wan2.1 (14B) & 13.07 & 0.29 & 0.52 & 0.40 & 0.34 \\
Motion Canvas~\cite{xing2025motioncanvas} & I2V & Wan2.1 (14B) & 15.27 & 0.36 & 0.45 & 0.43 & 0.41 \\
\hline[dashed]
HunyuanCustom~\cite{hunyuancus} & V2V + inpaint & Hunyuan & \SetCell{bg=best} 21.50 & \SetCell{bg=second} 0.67 & 0.35 & 0.44 & 0.42 \\
VACE~\cite{vace} & V2V + inpaint & Wan2.1 (14B) & 19.89 & 0.65 & \SetCell{bg=second} 0.24 & \SetCell{bg=second} 0.48 & 0.45 \\
GenCompositor~\cite{gencompositor} & V2V + inpaint & CogVideoX & 18.69 & 0.62 & 0.31 & \SetCell{bg=second} 0.48 & \SetCell{bg=best} 0.50 \\
PISCO~\cite{pisco} & V2V + inpaint & CogVideoX & 19.33 & 0.57 & 0.42 & 0.43 & 0.42 \\
\method (Ours) & V2V & Wan2.1 (14B) & \SetCell{bg=second} 20.48 & \SetCell{bg=best} 0.71 & \SetCell{bg=best} 0.19 & \SetCell{bg=best} 0.49 & \SetCell{bg=second} 0.48 \\
\bottomrule
\end{tblr}
\end{adjustbox}
\end{table*}


\begin{table*}[t!]
\centering
\caption{\textbf{Quantitative evaluation of the Full Pipeline}. 
We evaluate the video quality, similarity to the input, and box alignment of our \method, and compare it with other baselines in \method benchmark. 
The \colorbox{best}{best} and \colorbox{second}{second-best} scores are highlighted.}
\label{tab:full_pipeline}

\begin{adjustbox}{width=0.95\linewidth}
\begin{tblr}{
  width=\linewidth,
  colspec={@{}X[2.5,l] c c X[1,c] X[1,c] X[1,c] X[1.2,c] X[1,c]@{}},
  colsep=2pt,
  stretch=0.75,        
  rows = {rowsep=1pt}, 
  row{1-Z} = {font=\small},
}
\toprule
\SetCell[r=2]{c} Model & 
\SetCell[r=2]{c} Type & 
\SetCell[r=2]{c} Base Model & 
\SetCell[c=3]{c} Similarity and Quality & & & 
\SetCell[c=2]{c} Box Alignment \\
\cmidrule[lr]{4-6} \cmidrule[lr]{7-8}
& & & PSNR ($\uparrow$) & SSIM ($\uparrow$) & LPIPS ($\downarrow$) & Tube IoU ($\uparrow$) & mAP ($\uparrow$) \\
\midrule
MagicMotion~\cite{magic} & I2V & CogVideoX 
& 1492.56 & 46.91 & 0.0065 
& \SetCell{bg=best} 0.73 & \SetCell{bg=best} 0.89 \\

Wan-Move~\cite{wan-move} & I2V & Wan2.1 (14B) 
& 977.80 & 41.22 &  \SetCell{bg=second} 0.0029 
& 0.53 & 0.57 \\

MotionCanvas~\cite{xing2025motioncanvas} & I2V & Wan2.1 (14B)  
& 957.10 & \SetCell{bg=second} 39.79 & 0.0034 
& \SetCell{bg=second} 0.62 & \SetCell{bg=second} 0.69 \\

\hline[dashed]

HunyuanCustom~\cite{hunyuancus} & V2V + inpaint & Hunyuan 
& 780.13 & 41.81 & 0.0035 
& 0.53 & 0.54 \\

VACE~\cite{vace} & V2V + inpaint & Wan2.1 (14B)
& 916.37 & 60.86 & 0.0084 
& 0.55 & 0.59 \\

GenCompositor~\cite{gencompositor} & V2V + inpaint & CogVideoX 
& 967.11 & 178.69 & 0.0916 
& 0.53 & 0.59 \\

PISCO~\cite{pisco} & V2V + inpaint & CogVideoX   
& \SetCell{bg=second} 772.09 & 67.97 & 0.0136 
& 0.45 & 0.42 \\

\method (Ours) & V2V & Wan2.1 (14B) 
& \SetCell{bg=best} 614.94 & \SetCell{bg=best} 34.13 & \SetCell{bg=best} 0.0028 
& 0.57 & 0.61 \\

\bottomrule
\end{tblr}
\end{adjustbox}
\end{table*}

\begin{table}[t!]
\centering
\caption{
\textbf{Quantitative evaluation of Cross-View Motion Transformation Module.} 
Comparison of baseline methods across two tasks, box sequence from first-frame view to video view (f2v) and video view to first-frame view (v2f). 
We estimate camera pose and depth using MegaSAM~\cite{megasam} and DepthAnything-v3 (DA-v3)~\cite{depthanything3}, and warp the four box corners from the first frame to the corresponding video frame. 
The \colorbox{best}{best} and \colorbox{second}{second-best} scores are highlighted. Higher ($\uparrow$) is better for all metrics.
}
\label{tab:deep}

\begin{adjustbox}{width=0.99\linewidth}
\begin{tblr}{
  width=\linewidth,
  colspec={@{}X[4,l]X[1.1,c]X[1.1,c]X[1.1,c]X[1.1,c]@{}},
  colsep=2pt,
  stretch=0.75,
  rows = {rowsep=0.5pt},
  row{1-Z} = {font=\small},
}
\toprule
Model & IoU$_{\text{f2v}}$ & mAP$_{\text{f2v}}$ & IoU$_{\text{v2f}}$ & mAP$_{\text{v2f}}$ \\
\midrule
Interpolation & \SetCell{bg=second} 0.67 & \SetCell{bg=second} 0.72 & 0.67 & 0.72 \\

MegaSAM~\cite{megasam} warping 
& 0.63 & 0.56 & 0.67 & 0.48 \\

DA-v3~\cite{depthanything3} warping  
& \SetCell{bg=second} 0.73 & 0.67 & \SetCell{bg=second} 0.70 & 0.57 \\

\method (Ours Stage I) 
& \SetCell{bg=best} 0.80 & \SetCell{bg=best} 0.91 
& \SetCell{bg=best} 0.77 & \SetCell{bg=best} 0.85 \\
\bottomrule
\end{tblr}
\end{adjustbox}

\end{table}

\noindent\textbf{Baselines.} \noindent We adopt off-the-shelf 3D estimation models, MegaSAM~\cite{megasam} and DA3~\cite{depthanything3}, to estimate camera motion and per-frame depth from the input video, and use the estimated 3D information to perform \textbf{depth-based warping}, projecting the user’s first-frame boxes onto subsequent frames. We also compare against the \textbf{interpolation} baseline that directly applies the original first-frame bounding box to all frames without any transformation.

\noindent\textbf{Quantitative results.} 
As shown in Table~\ref{tab:deep}, our method significantly outperforms both 2D interpolation and 3D-warping baselines, achieving an $IoU_{f2v}$ of 0.80 and an $mAP_{f2v}$ of 0.91. Our model can learn to implicitly take into account the camera motion from the input video and adapt to dynamic scene changes better than the baseline approaches.
Consequently, our predicted paths are more accurate and closely match ground-truth motion, reaching an $mAP_{v2f}$ of 0.85—a substantial improvement over the 0.53 and 0.65 achieved by the MegaSAM and Depth-Anything baselines.
  

\noindent\textbf{Qualitative result.} 
Fig.~\ref{fig:stage1_baseline} shows a visual example of the box transformation results obtained from different methods.
We evaluate our cross-view motion transformation module against several baseline approaches to demonstrate its robustness in handling complex camera dynamics. As illustrated in our qualitative analysis, simple Interpolation of bounding boxes fails to account for the shifting perspective, resulting in an off-track trajectory that drifts to the right of the intended path. While 3D warping methods—utilizing MegaSAM \cite{megasam} for depth and camera pose estimation—attempt to model the scene geometry, they remain highly susceptible to noisy depth maps and imprecise pose estimates. This leads to unstable, "jittery" box trajectories that fail to reach the target destination. In a representative case study involving a fish moving toward a static \textcolor{red}{red coral} anchor, these baselines either ignore the viewpoint change or produce erratic jumps. In contrast, our proposed cross-view transformation generates a smooth and stable path, accurately delivering the fish box to the coral as intended while maintaining realistic scaling throughout the sequence.



\subsection{Video Re-Synthesis Module}
We evaluate video re-synthesis quality on the DAVIS benchmark~\cite{davis}, processed following the same procedure as in training (Section~\ref{sec:stage2}). 
Given an input video with the target object masked and the first frame as an identity reference, the model aims to re-synthesize the object while preserving temporal consistency and background information.
Generation Similarity is measured against the original ground-truth video using  SSIM~\cite{ssim}, PSNR, and LPIPS~\cite{lpip}.
To evaluate box alignment, we use mAP~\cite{map} at a 0.5 threshold and Tube IoU, which better captures dynamic motion by measuring global spatiotemporal overlap rather than static per-frame precision.


\noindent\textbf{Baselines.} We compare our method with two groups of baselines: (1) image-to-video (I2V) methods, including MagicMotion~\cite{magic}, Wan-Move~\cite{wan-move}, and Motion Canvas~\cite{xing2025motioncanvas}; and (2) state-of-the-art video-to-video (V2V) models: VACE~\cite{vace} , HunyuanCustom~\cite{hunyuancus}, and GenCompositor~\cite{gencompositor}.

\noindent\textbf{Quantitative results.} 
Table~\ref{tab:resynthesis_comparison} shows the generation performance of all methods.
We outperform all baselines in SSIM, LPIPS, and Tube IoU, while achieving second-best performance in PSNR and mAP. These results demonstrate that our method produces high-fidelity videos with reasonable box alignment.

\noindent\textbf{Qualitative result}.~\Fref{fig:baseline2} shows example video re-synthesis results from our experiment. The I2V baselines frequently hallucinates background content and  fails to preserve the goat’s appearance and color. HunyuanCustom introduces  small hallucinated goats in the background, while VACE fails to maintain the brown goat’s identity, generating a white goat and altering the background. 
 GenCompositor produces comparatively lower video quality, which further degrades object identity. In contrast, our method preserves both object identity 
 and scene consistency. Our outputs remain highly faithful to the input video, 
correctly generating the brown goat while maintaining a consistent background.

\textbf{}

\begin{table}[t!]
\centering
\caption{\textbf{Ablation study of Cross-View Motion Transformation}. 
We demonstrate how different conditions affect the model's performance. While full trajectories: incorporates motion paths for both dynamic objects and background regions to capture global scene flow ; Box concatenation: appends initial reference boxes to the video sequence without adding noise, providing stable spatial anchors for the generation process; 2 task learning: combines "First-to-Video" and "Video-to-First" training into a single unified model.
The \colorbox{best}{best} and \colorbox{second}{second-best} scores are highlighted.}
\label{tab:ablation_stage1}

\begin{adjustbox}{width=0.95\linewidth}
\begin{tblr}{
  width=\linewidth,
  colspec={@{}X[1.5,l]X[2.3,c]X[2.3,c]X[2,c]X[2.5,c]X[2,c]X[1,c]@{}},
  colsep=2pt,
  stretch=0.75,
  rows={rowsep=0.5pt},
  row{1-Z}={font=\small},
}
\toprule
{} & {Full trajectories} & {w/o trajectories} & {w/o first frame} & {Box concatenation} & {2 task learning} & {Ours} \\
\midrule
IoU ($\uparrow$) 
& 0.66 
& 0.41 
& 0.32 
& \SetCell{bg=second} 0.67 
& \SetCell{bg=second} 0.67 
& \SetCell{bg=best} 0.80 \\

mAP ($\uparrow$)
& \SetCell{bg=second} 0.76 
& 0.33 
& 0.18 
& 0.73 
& 0.73 
& \SetCell{bg=best} 0.91 \\
\bottomrule
\end{tblr}
\end{adjustbox}
\end{table}

\begin{table}[t!]
\centering
\caption{\textbf{Ablation study of Video Re-synthesis Module}. 
We evaluate the impact of different training configurations on the model using our full integrated pipeline and custom dataset. 
The \colorbox{best}{best} and \colorbox{second}{second-best} scores are highlighted.}
\label{tab:ablation_stage2}

\begin{adjustbox}{width=0.95\linewidth}
\begin{tblr}{
  width=\linewidth,
  colspec={@{} X[c] X[c] | X[1.4, c] X[1.3, c] X[1.3, c] | X[1.5,c] X[1.5,c] @{}},
  colsep=2pt,
  stretch=0.75,
  rows={rowsep=0.5pt},
  row{1-Z}={font=\small},
}
\toprule
{Smooth\\Boxes} & {Data\\Aug} & FVD ($\downarrow$) & FID ($\downarrow$) & KID ($\downarrow$) & {Tube\\IoU ($\uparrow$)} & mAP ($\uparrow$) \\
\midrule
\xmark & \cmark 
& 785.00 & 39.14 & 0.0033 
& \SetCell{bg=best} 0.59 & \SetCell{bg=best} 0.63 \\

\cmark & \xmark 
& \SetCell{bg=second} 643.62 & \SetCell{bg=second} 37.91 & \SetCell{bg=second} 0.0030 
& 0.53 & 0.58 \\

\cmark & \cmark 
& \SetCell{bg=best} 614.94 & \SetCell{bg=best} 34.13 & \SetCell{bg=best} 0.0028 
& \SetCell{bg=second} 0.57 & \SetCell{bg=second} 0.61 \\
\bottomrule
\end{tblr}
\end{adjustbox}
\end{table}

\begin{figure*}[t]
    \centering
    \includegraphics[width=\textwidth]{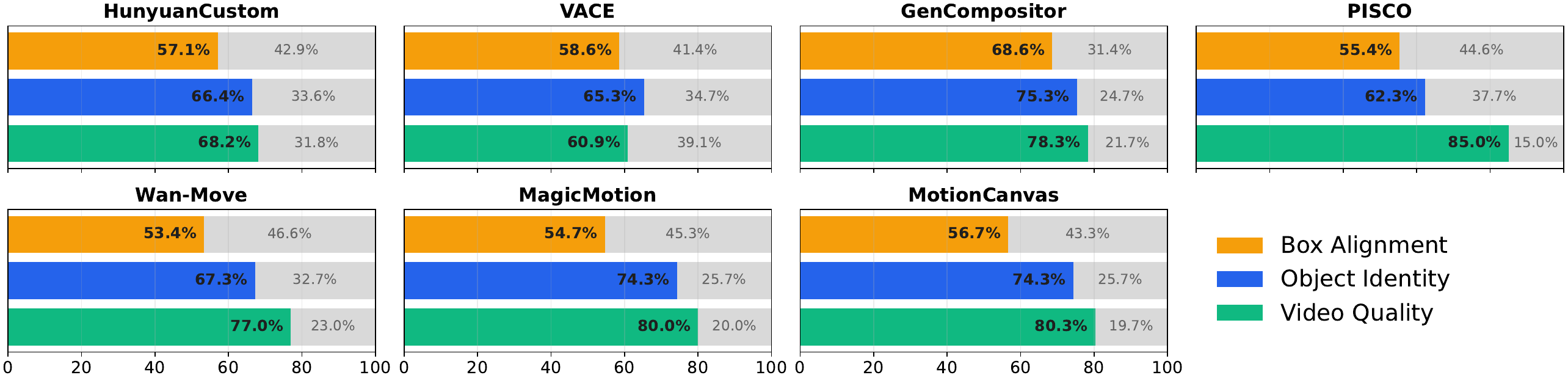}
    \vspace{-1.4em}
    \caption{
    \textbf{User study.} The colored segments represent the percentage of users who preferred our method over the respective baselines across three key dimensions.  }
    
    \label{fig:userstudy}
\end{figure*}

\subsection{Complete System Evaluation}
We evaluate the full framework (box transformation and video re-synthesis) on a manually curated test set of 70 videos. For each video, we annotate the target object location in the first and last frames on the first view to define a new motion path. These sparse annotations are interpolated into dense bounding boxes and used by the motion module to generate the full trajectory. The predicted trajectory is then used to re-synthesize the video.
We compare against video re-synthesis baselines, providing them with a shared inpainted video via Omnimatte~\cite{lee2025generative} since they lack native object removal capabilities.

We quantitatively assess video quality, control-following, and object identity preservation using FVD, FID, KID, and the VBench benchmark~\cite{huang2023vbench}
We also conduct a user study for subjective evaluation on the same 70 videos, in which participants make pairwise comparisons against baselines based on box alignment, object identity preservation, and video quality (details in the supplementary materials).

\noindent\textbf{Quantitative result}.
As shown in~\Tref{tab:full_pipeline}, our method outperforms all baselines in generative quality, achieving state-of-the-art scores in FVD, FID, and KID. While I2V models demonstrate higher Tube IoU and mAP—likely due to the greater freedom for hallucination in the I2V setup—our model remains highly competitive with other V2V baselines in box alignment. 
Furthermore, our method achieves the best results on VBench for Imaging Quality, Motion Smoothness, and Temporal Flickering while remaining competitive across all other metrics (see~\Tref{tab:vbench}).
Our user study (\Fref{fig:userstudy}) confirms these practical advantages. Users consistently preferred our method over all baselines in terms of video quality, object identity preservation, and motion controllability. 
Specifically, preference for our video quality exceeded 77\% against most baselines (peaking at 85.0\% against PISCO), while preference for object identity reached up to 75.3\% (against GenCompositor).

\noindent\textbf{Qualitative result}.
\Fref{fig:baseline_full} compares our full pipeline against baselines. MagicMotion fails to preserve the camera path, replacing the child with the mother and altering the background. 
While MotionCanvas preserves camera motion, it exhibits background drift and inconsistent identities for both subjects. 
HunyuanCustom and VACE suffer from identity loss and hallucinations, while GenCompositor shows poor video quality and perspective mismatches. 
In contrast, our method maintains high identity fidelity and background consistency, producing natural motion as the child realistically moves away from the woman.



\subsection{Ablation study}
We validate the design choices of the box transformation model in Table~\ref{tab:ablation_stage1}.
First, we confirmed the importance of our conditional inputs:
Removing key conditional inputs (either the point trajectory or the first-frame reference) causes a significant performance drop. Training with noisy box motion similarly degrades performance, underscoring the need for high-quality data. 
Second, we explored alternative task formulations (``inpainting-style'' setup and ``Combine 2 tasks'') and found that both alternatives underperformed, confirming the superiority of our chosen architecture.

For the Video Re-Synthesis model (\Tref{tab:ablation_stage2}), omitting box smoothing during training improves alignment (Tube IoU and mAP) but degrades generative quality (FID, FVD, KID), demonstrating its necessity for balancing control and visual fidelity. Additionally, removing data augmentation leads to performance drops across all metrics, underscoring its essential role in improving both video quality and box alignment (see supplementary materials for visual comparisons).

\section{Conclusion}
\label{sec:conclusion}

We present \method, a novel framework for editing object motion in videos. 
We introduced the setting of video editing with first-frame object motion design, in which users intuitively specify complex motion paths from a single anchor frame. 
To address the key challenge of cross-view spatial misalignment, our two-stage pipeline first employs a transformation module to predict a dense, frame-aligned bounding-box sequence from the user's first-frame path. 
A conditional re-synthesis model then uses this sequence to seamlessly ``erase-and-resynthesize'' the object, preserving the original video's content while making the selected object follow the user's new intended motion path. 
Experiments on real-world videos demonstrate high-quality, temporally consistent results, enabling new creative applications such as motion re-planning in post-production.

{
    \small
    \bibliographystyle{splncs04}
    \bibliography{main}

@String(CVPR= {IEEE Conf. Comput. Vis. Pattern Recog.})

@String(ICCV= {Int. Conf. Comput. Vis.})

@String(ECCV= {Eur. Conf. Comput. Vis.})

@String(ICLR = {Int. Conf. Learn. Represent.})

@String(CVPR  = {CVPR})

@String(ICCV  = {ICCV})

@String(ECCV  = {ECCV})

@String(ICLR  = {ICLR})

@article{trajectoryattn,
  title={Trajectory attention for fine-grained video motion control},
  author={Xiao, Zeqi and Ouyang, Wenqi and Zhou, Yifan and Yang, Shuai and Yang, Lei and Si, Jianlou and Pan, Xingang},
  journal={arXiv preprint arXiv:2411.19324},
  year={2024}
}

@inproceedings{magic,
  title={Magicmotion: Controllable video generation with dense-to-sparse trajectory guidance},
  author={Li, Quanhao and Xing, Zhen and Wang, Rui and Zhang, Hui and Dai, Qi and Wu, Zuxuan},
  booktitle={Proceedings of the IEEE/CVF International Conference on Computer Vision},
  pages={12112--12123},
  year={2025}
}

@inproceedings{levitor,
  title={Levitor: 3d trajectory oriented image-to-video synthesis},
  author={Wang, Hanlin and Ouyang, Hao and Wang, Qiuyu and Wang, Wen and Cheng, Ka Leong and Chen, Qifeng and Shen, Yujun and Wang, Limin},
  booktitle={Proceedings of the Computer Vision and Pattern Recognition Conference},
  pages={12490--12500},
  year={2025}
}

@article{disco,
  title={O-DisCo-Edit: Object Distortion Control for Unified Realistic Video Editing},
  author={Chen, Yuqing and Wang, Junjie and Liu, Lin and Chu, Ruihang and Zhang, Xiaopeng and Tian, Qi and Yang, Yujiu},
  journal={arXiv preprint arXiv:2509.01596},
  year={2025}
}

@article{truong2024local,
  title={Local part attention for image stylization with text prompt},
  author={Truong, Quoc-Truong and Nguyen, Vinh-Tiep and Nguyen, Lan-Phuong and Cao, Hung-Phu and Luu, Duc-Tuan},
  journal={Neural Computing and Applications},
  volume={36},
  number={34},
  pages={21859--21871},
  year={2024},
  publisher={Springer}
}

@misc{qwen,
      title={Qwen-Image Technical Report}, 
      author={Chenfei Wu and Jiahao Li and Jingren Zhou and Junyang Lin and Kaiyuan Gao and Kun Yan and Sheng-ming Yin and Shuai Bai and Xiao Xu and Yilei Chen and Yuxiang Chen and Zecheng Tang and Zekai Zhang and Zhengyi Wang and An Yang and Bowen Yu and Chen Cheng and Dayiheng Liu and Deqing Li and Hang Zhang and Hao Meng and Hu Wei and Jingyuan Ni and Kai Chen and Kuan Cao and Liang Peng and Lin Qu and Minggang Wu and Peng Wang and Shuting Yu and Tingkun Wen and Wensen Feng and Xiaoxiao Xu and Yi Wang and Yichang Zhang and Yongqiang Zhu and Yujia Wu and Yuxuan Cai and Zenan Liu},
      year={2025},
      eprint={2508.02324},
      archivePrefix={arXiv},
      primaryClass={cs.CV},
      url={https://arxiv.org/abs/2508.02324}, 
}

@INPROCEEDINGS{map,
  author={J. {Cartucho} and R. {Ventura} and M. {Veloso}},
  booktitle={2018 IEEE/RSJ International Conference on Intelligent Robots and Systems (IROS)}, 
  title={Robust Object Recognition Through Symbiotic Deep Learning In Mobile Robots}, 
  year={2018},
  pages={2336-2341},
}

@inproceedings{iou,
  title={Generalized intersection over union: A metric and a loss for bounding box regression},
  author={Rezatofighi, Hamid and Tsoi, Nathan and Gwak, JunYoung and Sadeghian, Amir and Reid, Ian and Savarese, Silvio},
  booktitle={Proceedings of the IEEE/CVF conference on computer vision and pattern recognition},
  pages={658--666},
  year={2019}
}

@article{fvd,
  title={Towards accurate generative models of video: A new metric \& challenges},
  author={Unterthiner, Thomas and Van Steenkiste, Sjoerd and Kurach, Karol and Marinier, Raphael and Michalski, Marcin and Gelly, Sylvain},
  journal={arXiv preprint arXiv:1812.01717},
  year={2018}
}

@article{ssim,
  title={Image quality assessment: from error visibility to structural similarity},
  author={Wang, Zhou and Bovik, Alan C and Sheikh, Hamid R and Simoncelli, Eero P},
  journal={IEEE transactions on image processing},
  volume={13},
  number={4},
  pages={600--612},
  year={2004},
  publisher={IEEE}
}

@inproceedings{lpip,
  title={The unreasonable effectiveness of deep features as a perceptual metric},
  author={Zhang, Richard and Isola, Phillip and Efros, Alexei A and Shechtman, Eli and Wang, Oliver},
  booktitle={Proceedings of the IEEE conference on computer vision and pattern recognition},
  pages={586--595},
  year={2018}
}

@article{pisco,
  title={PISCO: Precise Video Instance Insertion with Sparse Control},
  author={Gao, Xiangbo and Li, Renjie and Chen, Xinghao and Wu, Yuheng and Feng, Suofei and Yin, Qing and Tu, Zhengzhong},
  journal={arXiv preprint arXiv:2602.08277},
  year={2026}
}

@inproceedings{lee2025generative,
  title={Generative omnimatte: Learning to decompose video into layers},
  author={Lee, Yao-Chih and Lu, Erika and Rumbley, Sarah and Geyer, Michal and Huang, Jia-Bin and Dekel, Tali and Cole, Forrester},
  booktitle={Proceedings of the Computer Vision and Pattern Recognition Conference},
  pages={12522--12532},
  year={2025}
}

@inproceedings{davis,
  title={A benchmark dataset and evaluation methodology for video object segmentation},
  author={Perazzi, Federico and Pont-Tuset, Jordi and McWilliams, Brian and Van Gool, Luc and Gross, Markus and Sorkine-Hornung, Alexander},
  booktitle={Proceedings of the IEEE conference on computer vision and pattern recognition},
  pages={724--732},
  year={2016}
}

@article{wan-move,
  title={Wan-move: Motion-controllable video generation via latent trajectory guidance},
  author={Chu, Ruihang and He, Yefei and Chen, Zhekai and Zhang, Shiwei and Xu, Xiaogang and Xia, Bin and Wang, Dingdong and Yi, Hongwei and Liu, Xihui and Zhao, Hengshuang and others},
  journal={arXiv preprint arXiv:2512.08765},
  year={2025}
}

@inproceedings{xing2025motioncanvas,
  title={Motioncanvas: Cinematic shot design with controllable image-to-video generation},
  author={Xing, Jinbo and Mai, Long and Ham, Cusuh and Huang, Jiahui and Mahapatra, Aniruddha and Fu, Chi-Wing and Wong, Tien-Tsin and Liu, Feng},
  booktitle={Proceedings of the Special Interest Group on Computer Graphics and Interactive Techniques Conference Conference Papers},
  pages={1--11},
  year={2025}
}

@article{gu2024via,
  title={VIA: Unified Spatiotemporal Video Adaptation Framework for Global and Local Video Editing},
  author={Gu, Jing and Fang, Yuwei and Skorokhodov, Ivan and Wonka, Peter and Du, Xinya and Tulyakov, Sergey and Wang, Xin Eric},
  journal={arXiv preprint arXiv:2406.12831},
  year={2024}
}

@article{xu2025freevis,
  title={FreeViS: Training-free Video Stylization with Inconsistent References},
  author={Xu, Jiacong and Mei, Yiqun and Zhang, Ke and Patel, Vishal M},
  journal={arXiv preprint arXiv:2510.01686},
  year={2025}
}

@InProceedings{genprop,
    author    = {Liu, Shaoteng and Wang, Tianyu and Wang, Jui-Hsien and Liu, Qing and Zhang, Zhifei and Lee, Joon-Young and Li, Yijun and Yu, Bei and Lin, Zhe and Kim, Soo Ye and Jia, Jiaya},
    title     = {Generative Video Propagation},
    booktitle = {Proceedings of the IEEE/CVF Conference on Computer Vision and Pattern Recognition (CVPR)},
    month     = {June},
    year      = {2025},
    pages     = {17712-17722}
}

@inproceedings{videopainter,
  title={Videopainter: Any-length video inpainting and editing with plug-and-play context control},
  author={Bian, Yuxuan and Zhang, Zhaoyang and Ju, Xuan and Cao, Mingdeng and Xie, Liangbin and Shan, Ying and Xu, Qiang},
  booktitle={Proceedings of the Special Interest Group on Computer Graphics and Interactive Techniques Conference Conference Papers},
  pages={1--12},
  year={2025}
}

@article{depthanything3,
  title={Depth Anything 3: recovering the visual space from any views},
  author={Haotong Lin and Sili Chen and Jun Hao Liew and Donny Y. Chen and Zhenyu Li and Guang Shi and Jiashi Feng and Bingyi Kang},
  journal={arXiv preprint arXiv:2511.10647},
  year={2025}
}

@article{gencompositor,
  title={Gencompositor: generative video compositing with diffusion transformer},
  author={Yang, Shuzhou and Li, Xiaoyu and Cun, Xiaodong and Wang, Guangzhi and Li, Lingen and Shan, Ying and Zhang, Jian},
  journal={arXiv preprint arXiv:2509.02460},
  year={2025}
}

@InProceedings{huang2023vbench,
      title={{VBench}: Comprehensive Benchmark Suite for Video Generative Models},
      author={Huang, Ziqi and He, Yinan and Yu, Jiashuo and Zhang, Fan and Si, Chenyang and Jiang, Yuming and Zhang, Yuanhan and Wu, Tianxing and Jin, Qingyang and Chanpaisit, Nattapol and Wang, Yaohui and Chen, Xinyuan and Wang, Limin and Lin, Dahua and Qiao, Yu and Liu, Ziwei},
      booktitle={Proceedings of the IEEE/CVF Conference on Computer Vision and Pattern Recognition},
      year={2024}
}

@inproceedings{vace,
    title = {VACE: All-in-One Video Creation and Editing},
    author = {Jiang, Zeyinzi and Han, Zhen and Mao, Chaojie and Zhang, Jingfeng and Pan, Yulin and Liu, Yu},
    booktitle = {Proceedings of the IEEE/CVF International Conference on Computer Vision},
    pages = {17191-17202},
    year = {2025}
}

@article{revideo,
  title={Revideo: Remake a video with motion and content control},
  author={Mou, Chong and Cao, Mingdeng and Wang, Xintao and Zhang, Zhaoyang and Shan, Ying and Zhang, Jian},
  journal={Advances in Neural Information Processing Systems},
  volume={37},
  pages={18481--18505},
  year={2024}
}

@inproceedings{karaev23cotracker,
  title     = {CoTracker: It is Better to Track Together},
  author    = {Nikita Karaev and Ignacio Rocco and Benjamin Graham and Natalia Neverova and Andrea Vedaldi and Christian Rupprecht},
  booktitle = {Proc. {ECCV}},
  year      = {2024}
}

@inproceedings{ge2023preserve,
  title={Preserve your own correlation: A noise prior for video diffusion models},
  author={Ge, Songwei and Nah, Seungjun and Liu, Guilin and Poon, Tyler and Tao, Andrew and Catanzaro, Bryan and Jacobs, David and Huang, Jia-Bin and Liu, Ming-Yu and Balaji, Yogesh},
  booktitle={Proceedings of the IEEE/CVF International Conference on Computer Vision},
  pages={22930--22941},
  year={2023}
}

@article{recammaster,
  title={ReCamMaster: Camera-Controlled Generative Rendering from A Single Video},
  author={Bai, Jianhong and Xia, Menghan and Fu, Xiao and Wang, Xintao and Mu, Lianrui and Cao, Jinwen and Liu, Zuozhu and Hu, Haoji and Bai, Xiang and Wan, Pengfei and others},
  journal={arXiv preprint arXiv:2503.11647},
  year={2025}
}

@inproceedings{megasam,
  title     = {{MegaSaM}: Accurate, Fast and Robust Structure and Motion from Casual Dynamic Videos},
  author    = {Li, Zhengqi and Tucker, Richard and Cole, Forrester and Wang, Qianqian and Jin, Linyi and Ye, Vickie and Kanazawa, Angjoo and Holynski, Aleksander and Snavely, Noah},
  booktitle = {Proceedings of the IEEE/CVF Conference on Computer Vision and Pattern Recognition},
  year      = {2025}}

@article{shapemotion,
  title={Shape-for-Motion: Precise and Consistent Video Editing with 3D Proxy},
  author={Liu, Yuhao and Wang, Tengfei and Liu, Fang and Wang, Zhenwei and Lau, Rynson WH},
  journal={arXiv preprint arXiv:2506.22432},
  year={2025}
}

@inproceedings{deva,
  title={Tracking Anything with Decoupled Video Segmentation},
  author={Cheng, Ho Kei and Oh, Seoung Wug and Price, Brian and Schwing, Alexander and Lee, Joon-Young},
  booktitle={ICCV},
  year={2023}
}

@misc{hunyuancus,
      title={HunyuanCustom: A Multimodal-Driven Architecture for Customized Video Generation}, 
      author={Teng Hu and Zhentao Yu and Zhengguang Zhou and Sen Liang and Yuan Zhou and Qin Lin and Qinglin Lu},
      year={2025},
      eprint={2505.04512},
      archivePrefix={arXiv},
      primaryClass={cs.CV},
      url={https://arxiv.org/abs/2505.04512}, 
}

@article{wan2025,
      title={Wan: Open and Advanced Large-Scale Video Generative Models}, 
      author={Team Wan and Ang Wang and Baole Ai and Bin Wen and Chaojie Mao and Chen-Wei Xie and Di Chen and Feiwu Yu and Haiming Zhao and Jianxiao Yang and Jianyuan Zeng and Jiayu Wang and Jingfeng Zhang and Jingren Zhou and Jinkai Wang and Jixuan Chen and Kai Zhu and Kang Zhao and Keyu Yan and Lianghua Huang and Mengyang Feng and Ningyi Zhang and Pandeng Li and Pingyu Wu and Ruihang Chu and Ruili Feng and Shiwei Zhang and Siyang Sun and Tao Fang and Tianxing Wang and Tianyi Gui and Tingyu Weng and Tong Shen and Wei Lin and Wei Wang and Wei Wang and Wenmeng Zhou and Wente Wang and Wenting Shen and Wenyuan Yu and Xianzhong Shi and Xiaoming Huang and Xin Xu and Yan Kou and Yangyu Lv and Yifei Li and Yijing Liu and Yiming Wang and Yingya Zhang and Yitong Huang and Yong Li and You Wu and Yu Liu and Yulin Pan and Yun Zheng and Yuntao Hong and Yupeng Shi and Yutong Feng and Zeyinzi Jiang and Zhen Han and Zhi-Fan Wu and Ziyu Liu},
      journal = {arXiv preprint arXiv:2503.20314},
      year={2025}
}

@article{cogvideox,
  title={CogVideoX: Text-to-Video Diffusion Models with An Expert Transformer},
  author={Yang, Zhuoyi and Teng, Jiayan and Zheng, Wendi and Ding, Ming and Huang, Shiyu and Xu, Jiazheng and Yang, Yuanming and Hong, Wenyi and Zhang, Xiaohan and Feng, Guanyu and others},
  journal={arXiv preprint arXiv:2408.06072},
  year={2024}
}

@article{makevideo,
  title={Make-a-video: Text-to-video generation without text-video data},
  author={Singer, Uriel and Polyak, Adam and Hayes, Thomas and Yin, Xi and An, Jie and Zhang, Songyang and Hu, Qiyuan and Yang, Harry and Ashual, Oron and Gafni, Oran and others},
  journal={arXiv preprint arXiv:2209.14792},
  year={2022}
}

@article{moviegen,
  title={Movie gen: A cast of media foundation models},
  author={Polyak, Adam and Zohar, Amit and Brown, Andrew and Tjandra, Andros and Sinha, Animesh and Lee, Ann and Vyas, Apoorv and Shi, Bowen and Ma, Chih-Yao and Chuang, Ching-Yao and others},
  journal={arXiv preprint arXiv:2410.13720},
  year={2024}
}

@inproceedings{qi2023fatezero,
  title={Fatezero: Fusing attentions for zero-shot text-based video editing},
  author={Qi, Chenyang and Cun, Xiaodong and Zhang, Yong and Lei, Chenyang and Wang, Xintao and Shan, Ying and Chen, Qifeng},
  booktitle={Proceedings of the IEEE/CVF International Conference on Computer Vision},
  pages={15932--15942},
  year={2023}
}

@article{schneider2025neural,
  title={Neural Atlas Graphs for Dynamic Scene Decomposition and Editing},
  author={Schneider, Jan Philipp and Bisht, Pratik Singh and Chugunov, Ilya and Kolb, Andreas and Moeller, Michael and Heide, Felix},
  journal={arXiv preprint arXiv:2509.16336},
  year={2025}
}

@inproceedings{bai2025uniedit,
  title={Uniedit: A unified tuning-free framework for video motion and appearance editing},
  author={Bai, Jianhong and He, Tianyu and Wang, Yuchi and Guo, Junliang and Hu, Haoji and Liu, Zuozhu and Bian, Jiang},
  booktitle={Proceedings of the 33rd ACM International Conference on Multimedia},
  pages={10171--10180},
  year={2025}
}

@inproceedings{chung2024style,
  title={Style injection in diffusion: A training-free approach for adapting large-scale diffusion models for style transfer},
  author={Chung, Jiwoo and Hyun, Sangeek and Heo, Jae-Pil},
  booktitle={Proceedings of the IEEE/CVF conference on computer vision and pattern recognition},
  pages={8795--8805},
  year={2024}
}

@inproceedings{yang2023rerender,
  title={Rerender a video: Zero-shot text-guided video-to-video translation},
  author={Yang, Shuai and Zhou, Yifan and Liu, Ziwei and Loy, Chen Change},
  booktitle={SIGGRAPH Asia 2023 Conference Papers},
  pages={1--11},
  year={2023}
}

@article{wang2023zero,
  title={Zero-shot video editing using off-the-shelf image diffusion models},
  author={Wang, Wen and Jiang, Yan and Xie, Kangyang and Liu, Zide and Chen, Hao and Cao, Yue and Wang, Xinlong and Shen, Chunhua},
  journal={arXiv preprint arXiv:2303.17599},
  year={2023}
}

@article{wang2024boximator,
  title={Boximator: Generating rich and controllable motions for video synthesis},
  author={Wang, Jiawei and Zhang, Yuchen and Zou, Jiaxin and Zeng, Yan and Wei, Guoqiang and Yuan, Liping and Li, Hang},
  journal={arXiv preprint arXiv:2402.01566},
  year={2024}
}

@inproceedings{wiles2020synsin,
  title={Synsin: End-to-end view synthesis from a single image},
  author={Wiles, Olivia and Gkioxari, Georgia and Szeliski, Richard and Johnson, Justin},
  booktitle={Proceedings of the IEEE/CVF conference on computer vision and pattern recognition},
  pages={7467--7477},
  year={2020}
}

@inproceedings{huang2025fine,
  title={Fine-grained controllable video generation via object appearance and context},
  author={Huang, Hsin-Ping and Su, Yu-Chuan and Sun, Deqing and Jiang, Lu and Jia, Xuhui and Zhu, Yukun and Yang, Ming-Hsuan},
  booktitle={2025 IEEE/CVF Winter Conference on Applications of Computer Vision (WACV)},
  pages={3698--3708},
  year={2025},
  organization={IEEE}
}

@article{li2023trackdiffusion,
  title={Trackdiffusion: Tracklet-conditioned video generation via diffusion models},
  author={Li, Pengxiang and Chen, Kai and Liu, Zhili and Gao, Ruiyuan and Hong, Lanqing and Zhou, Guo and Yao, Hua and Yeung, Dit-Yan and Lu, Huchuan and Jia, Xu},
  journal={arXiv preprint arXiv:2312.00651},
  year={2023}
}

@inproceedings{wang2024motionctrl,
  title={Motionctrl: A unified and flexible motion controller for video generation},
  author={Wang, Zhouxia and Yuan, Ziyang and Wang, Xintao and Li, Yaowei and Chen, Tianshui and Xia, Menghan and Luo, Ping and Shan, Ying},
  booktitle={ACM SIGGRAPH 2024 Conference Papers},
  pages={1--11},
  year={2024}
}

@article{qiu2024freetraj,
  title={Freetraj: Tuning-free trajectory control in video diffusion models},
  author={Qiu, Haonan and Chen, Zhaoxi and Wang, Zhouxia and He, Yingqing and Xia, Menghan and Liu, Ziwei},
  journal={arXiv preprint arXiv:2406.16863},
  year={2024}
}

@article{mou2024revideo,
  title={Revideo: Remake a video with motion and content control},
  author={Mou, Chong and Cao, Mingdeng and Wang, Xintao and Zhang, Zhaoyang and Shan, Ying and Zhang, Jian},
  journal={Advances in Neural Information Processing Systems},
  volume={37},
  pages={18481--18505},
  year={2024}
}

@misc{videocrafter2,
      title={VideoCrafter2: Overcoming Data Limitations for High-Quality Video Diffusion Models}, 
      author={Haoxin Chen and Yong Zhang and Xiaodong Cun and Menghan Xia and Xintao Wang and Chao Weng and Ying Shan},
      year={2024},
      eprint={2401.09047},
      archivePrefix={arXiv},
      primaryClass={cs.CV}
}

@misc{videocrafter1,
      title={VideoCrafter1: Open Diffusion Models for High-Quality Video Generation}, 
      author={Haoxin Chen and Menghan Xia and Yingqing He and Yong Zhang and Xiaodong Cun and Shaoshu Yang and Jinbo Xing and Yaofang Liu and Qifeng Chen and Xintao Wang and Chao Weng and Ying Shan},
      year={2023},
      eprint={2310.19512},
      archivePrefix={arXiv},
      primaryClass={cs.CV}
}

@article{xing2023dynamicrafter,
      title={DynamiCrafter: Animating Open-domain Images with Video Diffusion Priors}, 
      author={Jinbo Xing and Menghan Xia and Yong Zhang and Haoxin Chen and Xintao Wang and Tien-Tsin Wong and Ying Shan},
      year={2023},
      eprint={2310.12190},
      archivePrefix={arXiv},
      primaryClass={cs.CV}
}

@article{lvdm,
      title={Latent Video Diffusion Models for High-Fidelity Long Video Generation}, 
      author={Yingqing He and Tianyu Yang and Yong Zhang and Ying Shan and Qifeng Chen},
      year={2022},
      eprint={2211.13221},
      archivePrefix={arXiv},
      primaryClass={cs.CV}
}

@article{hu2022lora,
  title={Lora: Low-rank adaptation of large language models.},
  author={Hu, Edward J and Shen, Yelong and Wallis, Phillip and Allen-Zhu, Zeyuan and Li, Yuanzhi and Wang, Shean and Wang, Liang and Chen, Weizhu and others},
  journal={Iclr},
  volume={1},
  number={2},
  pages={3},
  year={2022}
}
}

\clearpage
\appendix
\section{Additional applications}


Beyond the main application of altering object motion, we also extend our framework to support several additional tasks. 
\\
\noindent\textbf{In-place object editing :} In particular, users can edit the object in the first frame (\eg, modifying its appearance or attributes) and our system will propagate the edited object consistently across the rest of the video. This enables flexible object-level video editing while maintaining control over the object's trajectory (examples are shown on the webpage).
\\
\noindent\textbf{Object retining:} Furthermore, our method allows for retiming the object's motion by manipulating its location in the first-frame view. Users can make the object move faster, slower, or even remain stationary. Keeping the object stationary with our framework is very intuitive and easy for the user. For instance, to freeze the object, one simply sets the object’s bounding box in the last frame to be identical to that in the first frame, in the first-frame path design. Stage 1 of our method (\method) then automatically transforms the bounding boxes in the first-frame path design to video view, accounting for the camera motion in the video. Additional examples are provided on the project webpage.




\section{Training details}

\textbf{Cross-View Motion Transformation:}
We train the model from scratch using 1 NVIDIA H100 node. The training resolution is set to $488 \times 832$  with 73 frames per clip, and we use 28 inference steps during evaluation.

\noindent \textbf{Video Resynthesis:}
We fine-tune the Wan 2.1 14B base model using LoRA on 4 NVIDIA H100 nodes. Training is conducted at a resolution of  $488 \times 832$ with 81-frame video clips. During fine-tuning, LoRA fine-tuning are applied only to the DiT blocks, while the rest of the base model remains frozen.

For the conditional inputs, we use three video embeddings: the masked video, the object mask, and the inpainting mask. The embedding networks for these inputs are trained with full parameter updates, since they are initialized from scratch rather than loaded from the pretrained model.

\section{User Study}

The user study design is illustrated in~\Fref{fig:userstudy}. We conducted a comprehensive pairwise comparison study to evaluate our method against baseline approaches.
\\

\noindent\textbf{Study Setup.}
\\
\noindent\textit{\textbf{Inputs:}} The evaluation was conducted on a full set of 100 generated videos.
\\    
\noindent\textit{\textbf{Comparison Protocol:}} We adopted a pairwise comparison paradigm. In each trial, participants were presented with two videos: one generated by our method and one generated by a single baseline model.
\\
\noindent\textit{\textbf{Interface Design:}} The interface displayed the two videos side-by-side to facilitate direct comparison while avoiding visual redundancy or distracting elements. No additional cues were provided that could bias participants.
\\
\noindent\textit{\textbf{Participants and Responses:}} We recruited 46 participants, collecting a total of 1,656 responses across all evaluation criteria.
\\
\\
\noindent\textbf{Evaluation Criteria.}
To comprehensively evaluate object-centric video generation while minimizing ambiguity and cognitive bias, we decomposed the assessment into three independent criteria. Each criterion was evaluated through a separate question to ensure focused and reliable judgments.

\noindent\textit{\textbf{Controllability.}} Evaluates motion accuracy, i.e., how well the generated object's trajectory follows the user-specified input path.
    
\noindent\textit{\textbf{Object Identity.}} Measures identity preservation, assessing whether the generated object maintains consistent visual appearance throughout the video sequence.
    
\noindent\textit{\textbf{Video Quality.}} Assesses overall perceptual quality, including visual realism, temporal coherence, and the presence of artifacts.
\\
\\
\noindent\textbf{Results Reporting.}
The results of the pairwise comparisons are summarized in the main paper, Fig. 9 . We report performance using the preference percentage, defined as the proportion of responses in which participants selected our method over the baseline for a given criterion.

The preference ratio ($\mathcal{P}$) is computed as:
\[
\mathcal{P} =
\frac{\#\text{Responses Preferring Our Method}}
{\text{Total \#Responses}}
\times 100\%.
\]
\\
\\
\noindent\textbf{Overall Findings.}
Across all three evaluation criteria—\textbf{Box Alignment}, \textbf{Object Identity}, and \textbf{Video Quality}—participants demonstrate a clear and consistent preference for our method over the corresponding baselines. These results validate the effectiveness of our approach in improving motion controllability, identity preservation, and overall perceptual quality in object-centric video generation.

\begin{figure*}[t!]
    \centering
    \begin{subfigure}{0.48\linewidth}
        \centering
        \includegraphics[width=0.9\linewidth]{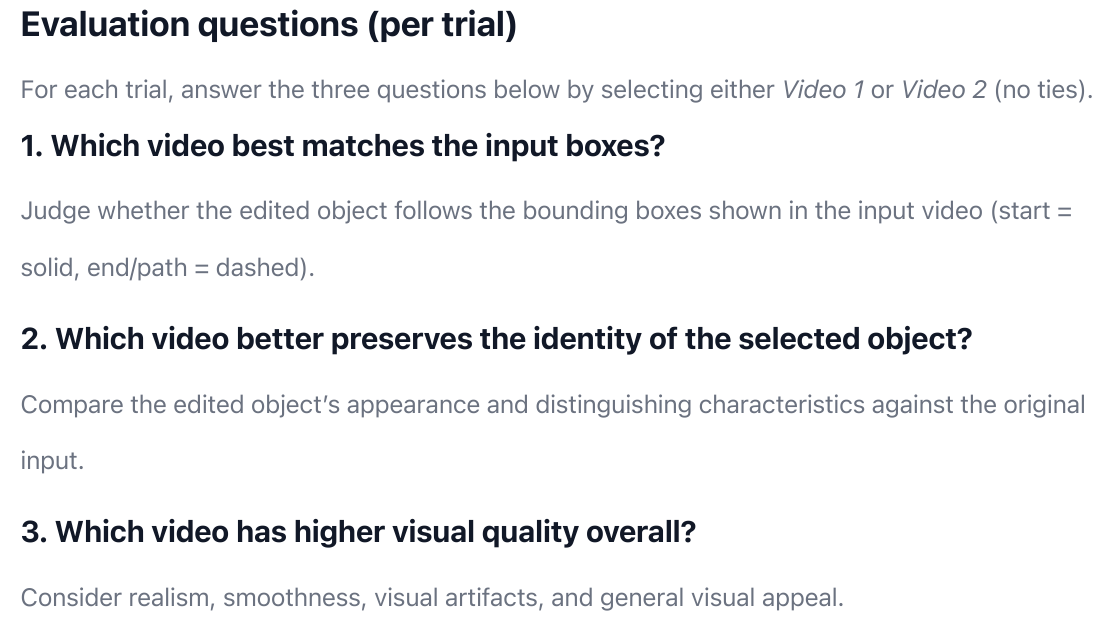}
        \caption{Explanation for questions }
        \label{fig:sub1}
    \end{subfigure}
    \hfill
    \begin{subfigure}{0.48\linewidth}
        \centering
        \includegraphics[width=0.9\linewidth]{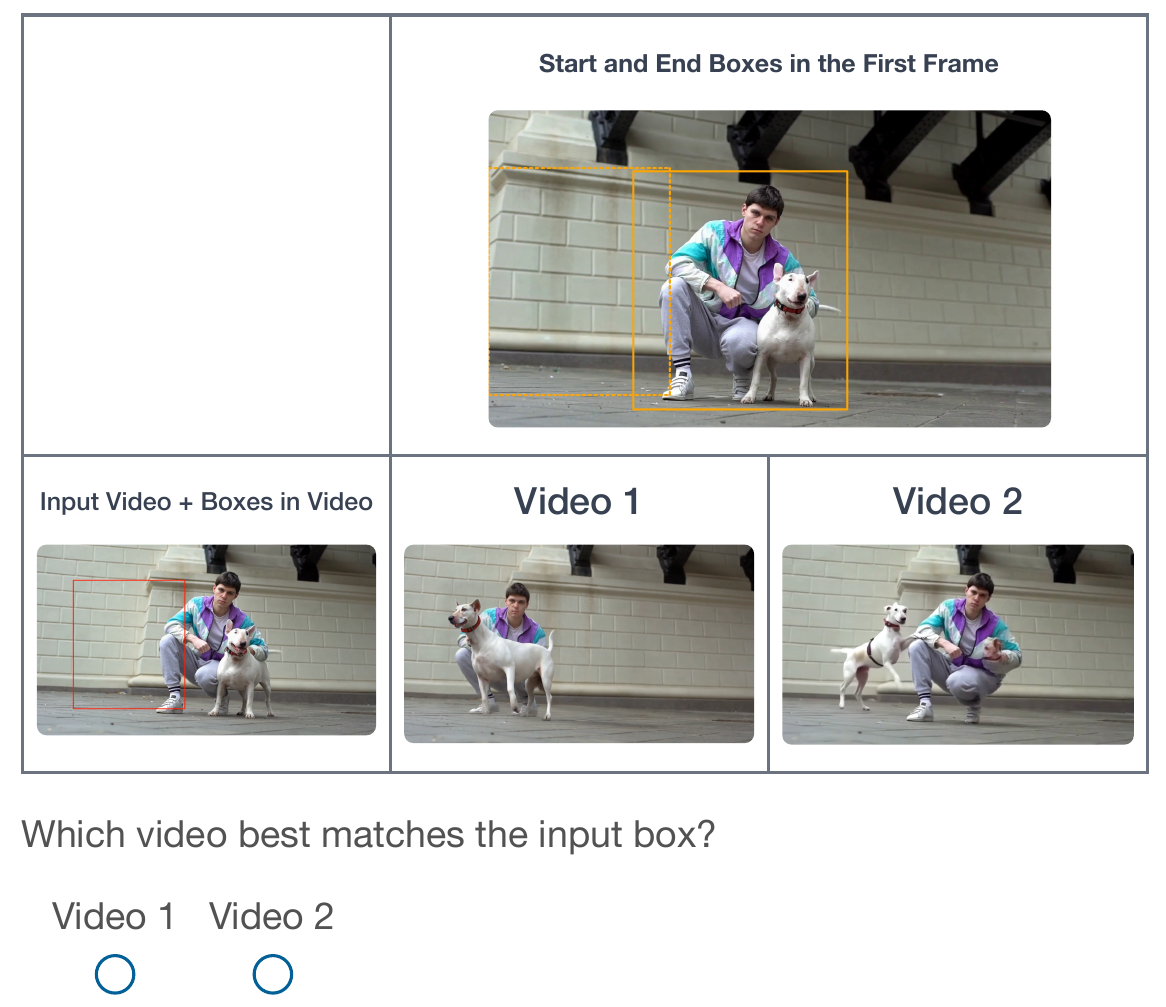}
        \caption{Question and layout for Controllability}
        \label{fig:sub2}
    \end{subfigure}

    \vspace{0.3cm}

    \begin{subfigure}{0.48\linewidth}
        \centering
        \includegraphics[width=0.9\linewidth]{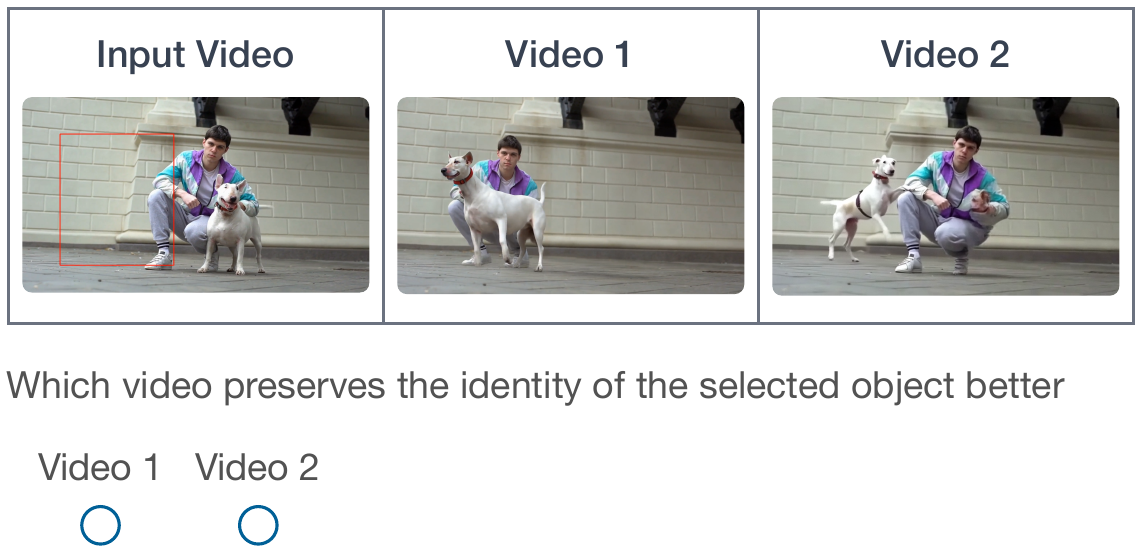}
        \caption{Question and layout for Object Identity}
        \label{fig:sub3}
    \end{subfigure}
    \hfill
    \begin{subfigure}{0.48\linewidth}
        \centering
        \includegraphics[width=0.9\linewidth]{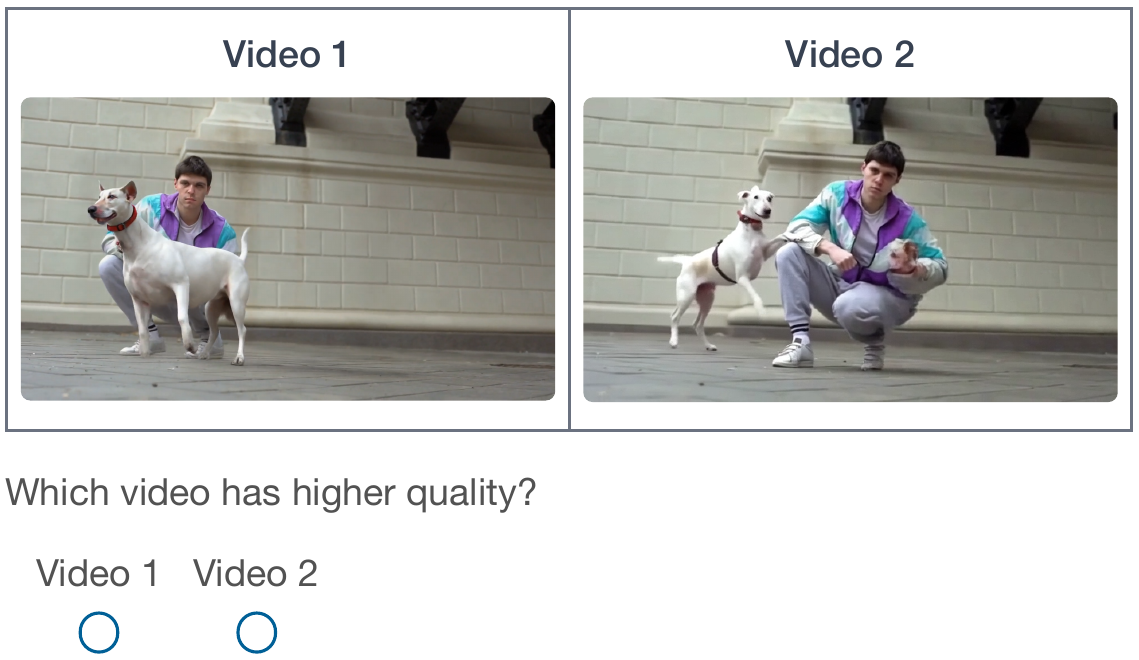}
        \caption{Question and layout for Video quality}
        \label{fig:sub4}
    \end{subfigure}

    \caption{\textbf{User study design.} Figure shows questions and comparison page shown to annotators to comapre the quality of our results with baselines.}
    \label{fig:userstudy}
\end{figure*}

\section{Additional cross-view motion transformation baselines }
We compare our, \method Stage I with MegaSAM~\cite{megasam} and DA3 (DepthAnything-v3~\cite{depthanything3}) baselines for cross-view motion transformation. These methods are used to estimate the camera parameters and per-frame depth. These are then used to warp the boxes from the first-frame view to each subsequent frame, for first-frame path to video-view transformation. Our method still outperforms these warping-based approaches, even when using different camera-calibration estimators. The results can be seen in~\Tref{tab:deep}. We use 2 different depths for warping the bounding boxes, i) \textit{center box:} the depth taken from the center of the box, ii)\textit{4 corners:} the depth of each of the coordinates of the bounding box.


\begin{table}[t!]
\centering
\caption{
\textbf{Cross-view motion transformation model compared with baselines.} 
Comparison of baseline methods from first-frame view to video view and vice versa. We use MegaSAM~\cite{megasam} and DepthAnything-v3 (DA-v3)~\cite{depthanything3} to estimate camera information and depth when warping the center path or the four box corners. 
The \colorbox{best}{best} and \colorbox{second}{second-best} scores are highlighted.
}
\label{tab:deep}

\begin{adjustbox}{width=0.95\linewidth}
\begin{tblr}{
  width=\linewidth,
  colspec={@{}X[6,l]X[1.3,c]X[1.3,c]X[1.3,c]X[1.3,c]@{}},
  stretch=0.75,
  colsep=2pt,
  rows={rowsep=0.5pt},
  row{1-Z}={font=\small},
}
\toprule
Model & IoU$_{\text{f2v}}$ & mAP$_{\text{f2v}}$ & IoU$_{\text{v2f}}$ & mAP$_{\text{v2f}}$ \\
\midrule
Interpolation
& 0.67 & 0.72 & 0.67 & 0.72 \\

MegaSAM warping (center)
& 0.69 & 0.64 & 0.71 & 0.53 \\

MegaSAM warping (corners)
& 0.63 & 0.56 & 0.67 & 0.48 \\

DA-v3 warping (center)
& \SetCell{bg=second} 0.79 & \SetCell{bg=second} 0.73 & \SetCell{bg=best} 0.83 & \SetCell{bg=second} 0.65 \\

DA-v3 warping (corners)
& 0.73 & 0.67 & \SetCell{bg=second} 0.77 & 0.57 \\

\method (Ours Stage I)
& \SetCell{bg=best} 0.80 & \SetCell{bg=best} 0.91 & \SetCell{bg=second} 0.77 & \SetCell{bg=best} 0.85 \\
\bottomrule
\end{tblr}
\end{adjustbox}

\end{table}

\section{Limitations and Future Work}



\begin{figure*}[t]
    \centering
    \includegraphics[width=0.9\textwidth]{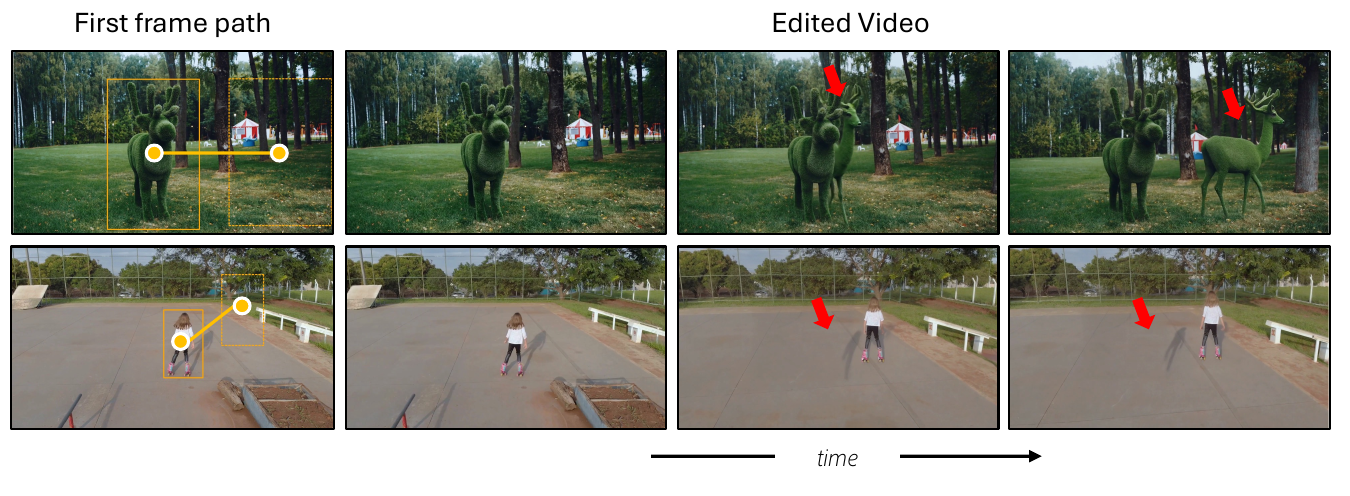}
    \vspace{-1.4em}
    \caption{\textbf{Limitation.} Figure shows two different cases where our method fails to generate desired results. \textbf{(top)} \method fails to inpaint the existing deer correctly, and instead creates a new deer following the trajectory. \textbf{(bottom)} \method can not inpaint effects, like shadows as they are not part of the inpainting bounding box of the skating girl.
    }
    
    \label{fig:limit}
\end{figure*}

The main limitations of our method are: i) In some instances, in some cases, the Stage II part of our model duplicates the object, which could be caused due to faliure of the inpainting part, as illustrated in the first example in~\Fref{fig:limit}. ii) In some cases involving shadows, reflections, or water effects, the framework can only move the object itself while failing to reproduce the associated effects, unless the inpainting bounding box covers the shadows also. This is not trivial to do on the user side. As a result, elements such as shadows or water ripples remain unchanged, as seen in the second example in~\Fref{fig:limit}. iii) In some rare cases, the Stage II model does not follow the object to the desired trajectory. iv) The cross-view box transformation (Stage I) model, although robust to different scales of object motion, can sometimes fail when the input view has extremely large camera motion.

\noindent\textbf{Future Work.}
We aim to further the quality of results by improving and correcting existing limitations mentioned in the above section. We believe the errors in inpainting and inability of the model to handle shadows or associated effects, limitations i) and ii), can be mitigates by training with paired inpainting data, involving shadows. We currently train our model with LoRA on Wan21.-14B. Training the model (all weights) with more data on a larger scale can improve the bounding box following. Lastly, to make the cross-view box transformation (Stage I) model more robust, we need to prepare synthetic camera view data (with large viewpoint changes) in unreal-engine. Presently we only train with Recammaster~\cite{recammaster} data and hence limited by their camera presents.








\end{document}